\documentclass[a4paper,fleqn]{cas-sc}

\usepackage[numbers]{natbib}

\usepackage{amsmath, amsfonts}
\usepackage{algorithm}
\usepackage{array}
\usepackage{graphicx}
\graphicspath{{images/}}
\usepackage{enumitem}
\usepackage{multicol}
\usepackage[dvipsnames]{xcolor}
\usepackage{multirow}
\usepackage{longtable}
\usepackage{lscape}
\usepackage{float}
\usepackage{tablefootnote}
\usepackage{tabularx}
\usepackage{booktabs}
\usepackage{adjustbox}
\usepackage{flushend}
\usepackage[noend]{algpseudocode}

\renewcommand{\textcolor}[3][]{#3}

\usepackage{placeins}
\usepackage{setspace}
\usepackage{soul}
\sethlcolor{yellow}

\begin{document}


    \title[mode = title]{Emission-Aware Reinforcement Learning for Sustainable Electric Vehicle Charging and Carbon Dioxide Reduction Under Varying Renewable Penetration}


    \author[1]{Ninglin Ou}
    [orcid=0009-0005-6796-0430]\ead{123113268@umail.ucc.ie}

    \author[2]{Mohammad A. Razzaque}
    [orcid=0000-0002-5572-057X]\ead{m.razzaque@tees.ac.uk}

    \author[1]{Iftekher Islam Shovon}
    [orcid=0000-0002-5678-4612] \ead{}

    \author[1]{Shafkat Khan Siam}
    [orcid=0000-0003-3324-6034] \ead{}

    \author[3]{Shafiuzzaman K Khadem}
    [orcid=0000-0001-5869-770X] \ead{shafi.khadem@ierc.ie}

    \author[1]{Krishnendu Guha}
    [] \ead{}

    \author[4,5]{Mayeen U Khandaker}
    [orcid=0000-0003-3772-294X] \ead{mayeenk@sunway.edu.my}

    \author[1]{Md. Noor-A-Rahim}
    [orcid=0000-0003-0587-3145] \ead{m.rahim@cs.ucc.ie}



    \affiliation[1]{organization={nasc Research, School of Computer Science \& IT, University College Cork}, country = {IE}}

    \affiliation[2]{organization={School of Computing, Engineering and Digital Technologies, Teesside University}, country={UK}}

    \affiliation[3]{organization={International Energy Research Centre, Tyndall National Institute, Cork}, country = {IE}}
    \affiliation[4]{organization={Applied Physics and Radiation Technologies Group, CCDCU, Faculty of Engineering and Technology, Sunway University}, country = {Malaysia}}
    \affiliation[5]{organization={Department of Physics, College of Science, Korea University}, country = {Republic of Korea}}

    \cortext[cor1]{Corresponding author: Md Noor-A-Rahim}



    \begin{abstract}
    The rapid growth of Electric Vehicle (EV) adoption challenges power distribution networks through peak load spikes, voltage instability, and transformer overloads from uncoordinated charging. While Model Predictive Control (MPC) and standard Reinforcement Learning (RL) methods have addressed these issues, existing approaches rarely treat real-time carbon intensity or fluctuating renewable energy (RE) availability as primary scheduling objectives, leaving substantial decarbonisation potential unrealised. This paper proposes an emission-aware RL strategy based on the Soft Actor-Critic (SAC) algorithm, with a multi-objective reward that penalises carbon emissions, curtailed on-site renewables, and unmet user demand. The agent is trained within a unified benchmarking framework on the EV2Gym platform, incorporating behind-the-meter solar and wind profiles, time-varying EirGrid carbon intensity data, and realistic workplace EV behaviour across 25 Electric Vehicle Supply Equipment (EVSE) units. Nine control strategies, including heuristics, emission-aware MPC variants, and the proposed RL agent, are compared under five renewable penetration scenarios (0\%–50\%) over ten independent runs each. The RL agent achieves a carbon intensity as low as 23.96 grams of carbon dioxide per kilowatt-hour (g$CO_2$/kWh) under 50\% wind penetration, representing up to 87\% emission reduction versus the uncontrolled baseline, and outperforms the external graph-based Power Distribution Network (PDN) benchmark (95–137 g$CO_2$/kWh). Transformer overload remains below 7 kWh across scenarios, against up to 1093 kWh for the As Fast As Possible (AFAP) heuristic, and renewable self-consumption reaches 52\% under combined wind and solar supply. Embedding carbon intensity forecasts into the RL state and reward aligns charging with low-emission periods while preserving grid compliance and user satisfaction.
    \end{abstract}




\maketitle

    \section{Introduction}
    \label{sec1}

    Electric vehicles (EVs) have rapidly transitioned from a niche to a global phenomenon, with worldwide EV sales surpassing 14 million in 2023, accounting for 25\% of new car sales in Europe~\cite{IEA2024}. The promise of reduced carbon emissions and decreased dependence on fossil fuels drive the electrification of transportation. However, the surge in EV adoption also poses significant challenges for power grids and the environment. Uncoordinated high-volume charging can cause steep peak loads, voltage instability, and transformer overloads, straining existing infrastructure. By 2050, the majority of distribution transformers could exceed their operational capacity due to the growth of unmanaged EV charging~\cite{rossi2025smart}. These issues require smarter charging strategies to avoid costly grid upgrades and maximise the environmental benefits of EVs. Indeed, intelligent scheduling of EV charging has been identified as essential to maintain power quality and reliability as EV penetration increases~\cite{panda2024multi}. The timing of charging directly affects the stability of the grid and the carbon footprint of transportation. Aligning charging with periods of abundant renewable energy (RE) generation can substantially reduce emissions~\cite{NationalEVAnnualReport}, underscoring the need to integrate RE considerations into charging control.

    Researchers have proposed a wide range of smart charging solutions to address these challenges. Classic optimisation approaches~\cite{csengor2018optimal}, including mixed-integer programming and Model Predictive Control (MPC)~\cite{holkar2010overview}, have been widely applied to schedule EV charging cost-effectively while ensuring compliance with network constraints. For example, several studies formulate charging strategies to flatten peak demand or minimise electricity costs for operators~\cite{panda2024multi}. Under demand response programs, optimal scheduling can improve load factors and reduce peak loads at parking facilities, even with uncertainty in EV arrivals~\cite{csengor2018optimal}. Beyond grid-centric objectives, some strategies incorporate vehicle-to-grid (V2G)~\cite{kempton2005vehicle} or vehicle-to-vehicle (V2V) energy exchange to enhance flexibility. V2G allows EVs to discharge power back into the grid during critical periods,  shaving peaks and providing ancillary services, while emerging V2V mechanisms enable energy trading among EVs to alleviate local charging bottlenecks~\cite{dan2024cooperative}.

    Reinforcement Learning (RL)~\cite{li2017deep} methods have gained increasing attention in EV charging control, owing to their ability to adapt to complex, uncertain environments and learn efficient scheduling policies. Recent studies show that RL-based controllers can reduce charging costs and improve renewable energy utilisation compared with rule-based or MPC strategies~\cite{9817043}. However, \textcolor{blue}{as summarised in Table~\ref{tab:literature_summary},} despite these advances, few studies explicitly target carbon emissions or incorporate real-time renewable availability into scheduling decisions. In most work, economic efficiency and grid reliability remain the primary objectives, while environmental impact is treated as secondary. This highlights the need for approaches that incorporate carbon intensity signals and varying levels of RE penetration into charging algorithms to achieve more sustainable outcomes.

    This paper addresses this gap by presenting a comprehensive simulation-based evaluation of EV charging scheduling strategies across a range of RE penetration scenarios. Leveraging the open-source EV2Gym platform~\cite{EV2Gym}, we implement and compare nine distinct control approaches, including simple heuristic policies, two MPC schemes, and an advanced RL strategy introduced in this work, within a common, standardised environment. Each strategy is evaluated against multiple performance criteria, including total carbon emissions, aggregate load profiles, transformer overloading incidents, and end-user charging satisfaction. \textcolor{blue}{This design enables a systematic examination of how both the control method and the level of renewable supply shape grid performance and environmental outcomes.}

    \textcolor{blue}{The novelty of this work does not lie in applying reinforcement learning to EV charging in itself, since RL has already been widely studied for charging coordination and smart grid energy management. Instead, the methodological contribution is threefold. First, EV charging is formulated here as an explicitly emission-aware control problem by incorporating time-varying grid carbon intensity and renewable availability directly into both the RL state representation and the reward design. Second, a multi-objective reward function is developed to balance carbon reduction, renewable utilisation, and user charging satisfaction, while discouraging excessive or operationally unrealistic discharge behaviour. Third, the proposed controller is evaluated within a realistic V2G-enabled and transformer-constrained benchmark and compared systematically against heuristic and MPC baselines across multiple renewable-penetration scenarios. Taken together, these elements shift the focus from generic smart charging to carbon-aware operational scheduling under practical grid conditions.}

    \begin{figure}
        \centering
        \includegraphics[width=1\linewidth]{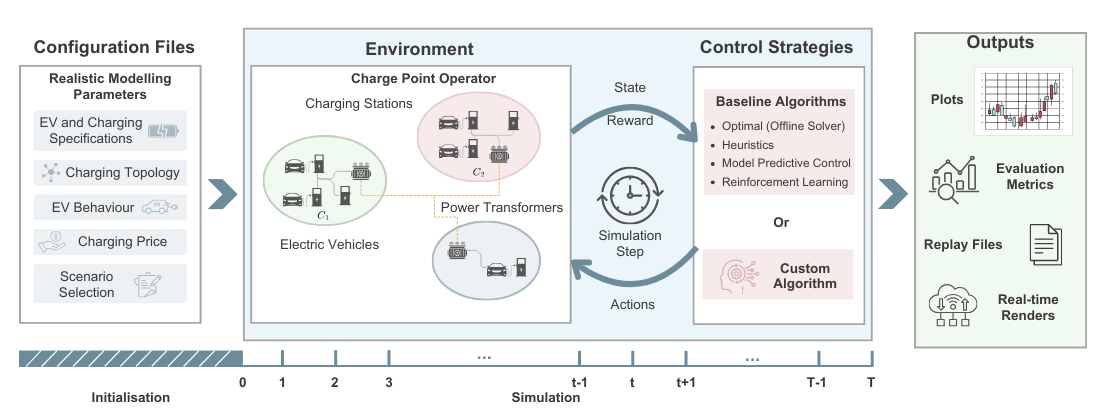}
        \caption{An EV2Gym simulation consists of three phases: configuration, where models are initialised; simulation, where system states evolve over $T$ steps under the chosen control algorithm; and evaluation, where performance metrics, replay files, and optional visualisations are generated~\cite{EV2Gym}.}
        \label{fig:EV2Gym_Architecture}
        \end{figure}

    The main contributions of this work are as follows.
    \begin{itemize}
        \item Development of a novel data-driven, emission-aware reinforcement learning (RL) agent-based EV charging strategy with a multi-objective reward function designed to balance grid compliance, user satisfaction, and carbon reduction.
       
        \item A systematic evaluation and comparison of diverse EV charging strategies under identical conditions, covering heuristic methods, MPC, and the proposed RL approach, highlighting their relative strengths and weaknesses.

        \item The explicit incorporation of real-time carbon intensity signals and varying renewable generation levels into the scheduling process, enabling a quantitative assessment of their impact on emissions and grid performance.  

        \item The development of a unified benchmarking framework within EV2Gym that integrates dynamic carbon signals, behind-the-meter solar and wind generation, and realistic EV behaviour, thereby ensuring reproducible and sustainability-oriented comparisons across control algorithms. 
    \end{itemize}

    \textcolor{blue}{The remainder of this paper is structured as follows. Section~\ref{related_work} reviews the most relevant literature on EV smart charging and carbon-aware control. Section~\ref{sec:methodology} presents the overall methodological framework. Section~\ref{simulator_modelling} details the EV2Gym simulator architecture and system modelling assumptions. Section~\ref{sec:energy_modelling} presents the formulation of energy flows and carbon emissions. Section~\ref{charging_strategy} introduces the proposed emission-aware charging strategy and baseline controllers, specifying the design of states, actions, and rewards. Section~\ref{results} reports and discusses the comparative results across control strategies and renewable penetration scenarios. Finally, Section~\ref{conclusion} concludes the paper and outlines directions for future research.}
    
    \begin{table*}[ht]
    \centering
    \caption{Summary of literature considering V2G support, renewable integration, carbon emissions, grid constraints, and control method classification}
    \label{tab:literature_summary}
    \begin{tabular}{>{\centering\arraybackslash}p{2.5cm}ccccc}
    \hline
    \textbf{Reference} & \textbf{V2G Support} & \textbf{Renewable Integration} & \textbf{Carbon Emissions} & \textbf{Grid Constraints} & \textbf{Control Method} \\
    \hline
    \cite{rossi2025smart}          & \checkmark & $\times$    & $\times$    & \checkmark & RL \\
    \cite{yilmaz2024reinforcement} & \checkmark & $\times$    & $\times$    & \checkmark & RL \\
    \cite{8710609}                 & \checkmark & \checkmark  & $\times$    & $\times$   & Optimisation \\
    \cite{qureshi2024multiobjective} & \checkmark & \checkmark & $\times$    & \checkmark & Optimisation \\
    \cite{panda2024multi}          & \checkmark & \checkmark  & $\times$    & \checkmark & Optimisation \\
    \cite{csengor2018optimal}      & \checkmark & \checkmark  & $\times$    & \checkmark & Optimisation \\
    \cite{hosseini2024optimizing}  & $\times$   & \checkmark  & $\times$    & $\times$   & AI / Heuristic \\
    \cite{kanchana2024optimizing}  & $\times$   & $\times$    & $\times$    & $\times$   & Optimisation \\
    \textbf{This study}            & \checkmark & \checkmark  & \checkmark  & \checkmark & RL \\
    \hline
    \end{tabular}
    \vspace{0.2cm}
    \footnotesize{\textbf{Legend:} \checkmark considered; $\times$ not considered.}
\end{table*}

    \section{Related Work}
    \label{related_work}
    Research on EV smart charging has explored optimisation methods, reinforcement learning, energy sharing mechanisms, and simulator platforms. This section reviews these approaches and their limitations, motivating the need for an emission-aware framework. A structured comparison of key representative studies is provided in Table~\ref{tab:literature_summary} to highlight their coverage of V2G, renewable integration, carbon emissions, grid constraints, and control methodologies.
    
    \subsection{Optimisation-Based Charging Strategies}
    Early smart charging methods frequently relied on optimisation and control techniques such as mixed-integer programming and Model Predictive Control (MPC) to minimise peak demand, reduce costs, or manage grid constraints~\cite{csengor2018optimal,9817043,DIAZLONDONO2024100326}. These approaches achieved high load factors and cost savings by forecasting EV behaviour and shifting charging to optimal time windows. Some studies extended these methods to incorporate uncertainties in photovoltaic (PV) generation or EV arrivals~\cite{9817043}, while others addressed heterogeneous charger and battery types through data-driven clustering~\cite{DIAZLONDONO2024100326}. Recent work has also employed stochastic convex optimisation to address multiple objectives in commercial charging stations~\cite{qureshi2024multiobjective}. Other studies explored heuristic and AI-driven approaches, for example, minimising charging time and travel distance~\cite{kanchana2024optimizing} or applying artificial intelligence mechanisms for smart transportation~\cite{hosseini2024optimizing}. \textcolor{blue}{Nevertheless, many of these models assume simplified linear battery behaviour, overlook degradation effects, and face scalability limitations when applied to large fleets or real-time operational settings.} Our framework addresses these limitations by supporting detailed two-stage battery charging curves, transformer-level constraints, and degradation-aware battery modelling, thereby enabling realistic and scalable benchmarking of optimisation strategies under practical grid conditions.

    \subsection{Reinforcement Learning for Smart Charging}
    RL has gained increasing attention in EV smart charging due to its ability to manage uncertainty, scalability, and dynamic environments. A recent review~\cite{QIU2023113052} outlines RL applications ranging from single-agent to multi-agent settings, targeting objectives such as cost minimisation, load balancing, and V2G integration. For example, Rossi et al.~\cite{rossi2025smart} demonstrated that RL agents can learn effective charging policies without requiring explicit forecasts.

    Despite this progress, many existing RL studies rely on simplified simulators that exclude critical constraints such as transformer capacity, nonlinear charging curves, and battery degradation~\cite{QIU2023113052}. Platforms such as ACN-Sim~\cite{8909765} and SustainGym~\cite{NEURIPS2023_ba748557} provide Gym-compatible environments but lack full V2G support and detailed grid modelling. Furthermore, Qiu et al.~\cite{qiu2024graph} introduced a graph-based RL model that jointly optimises EV routing and charging to reduce carbon emissions from charging, offering a valuable external benchmark for comparison. Our framework addresses these limitations by integrating realistic EV and grid models, supporting flexible RL development, and enabling reproducible benchmarking under standardised constraints.

    \textcolor{blue}{Recent studies have further advanced EV charging control toward grid-aware and carbon-aware formulations. Fan et al.~\cite{fan2024safety} proposed a safety-aware off-policy RL controller for EV charging station management in distribution networks, explicitly enforcing operational constraints under uncertainty in solar generation and electricity prices. Zhao et al.~\cite{zhao2025maasac} extended this line of work with a multi-agent asynchronous soft actor-critic framework for residential EV charging and discharging that jointly considers carbon intensity, charging anxiety, and transformer overload. In a related direction, Silva and Bessa~\cite{silva2025carbon} developed a forecast-driven carbon-aware tariff design using stochastic optimisation under uncertainty in renewable generation, demand, and grid carbon intensity. These studies highlight the growing interest in grid-aware and low-carbon charging control. However, they either emphasise safety or tariff design rather than direct emission-aware charger-level scheduling, or focus on residential coordination rather than a unified benchmark across heuristic, MPC, and RL methods under varying renewable penetration, as considered in this work.}

    \subsection{Vehicle-to-Grid and Energy Sharing}
    Vehicle-to-grid (V2G) and vehicle-to-vehicle (V2V) energy sharing can transform EV fleets into distributed energy resources, offering grid services and improving local energy efficiency. Koufakis et al.~\cite{8710609} proposed a hybrid MILP-based framework that coordinates V2V transfers to utilise surplus PV generation, achieving 12\% higher renewable self-consumption and 3.3\% cost savings. Meenakumar et al.~\cite{9242538} demonstrated that EV fleets can generate revenue via frequency regulation and peak shaving, although profitability is highly sensitive to pricing and battery wear. To address battery health concerns, Lee et al.~\cite{LEE20236624} incorporated degradation-aware dispatch in V2G optimisation to balance grid support and battery longevity.

    These studies highlight the potential of V2G and V2V but often rely on idealised assumptions such as perfect foresight and centralised coordination, while overlooking grid-level impacts. Our framework supports bi-directional charging with transformer-level constraints, calibrated battery ageing models, and heterogeneous user behaviour, enabling realistic evaluation of energy sharing strategies under operational limits and user variability.

    \subsection{Simulation Platforms for \textcolor{blue}{Electric Vehicle} Charging Research}
    A range of EV charging simulators has been developed to support algorithm design and evaluation, each with specific capabilities and limitations. Saxena~\cite{SAXENA2015720} introduced \textit{V2G-Sim}, one of the earliest tools for modelling EV battery dynamics and mobility patterns, but it is closed-source and lacks compatibility with modern RL-based controllers. \textit{EVLibSim} provides a flexible Java-based interface supporting V2G, battery swapping, and inductive charging. Yet, it omits distribution grid constraints and is not well-suited for integration with Python-based ML tools~\cite{RIGAS201899}. Building on this, \textit{EV-EcoSim} extends EV simulation with power system co-simulation, allowing voltage and transformer impact analysis. Still, it assumes homogeneous EV characteristics and does not support learning-based controllers. \textit{OPEN} enables co-simulation of EVs and other distributed energy resources with grid-aware modelling, but it relies on simplified EV assumptions and lacks Gym-style RL support~\cite{MORSTYN2020115397}.

    More recent platforms have improved interoperability and realism. \textit{ACN-Sim} provides empirical EV charging data, supports integration with power flow tools such as MATPOWER, and includes a Gym interface, but only models unidirectional (G2V) charging~\cite{8909765}. \textit{SustainGym} builds on ACN-Sim to enable standardised RL benchmarking with predefined observation, action, and reward spaces, but does not introduce V2G capabilities or richer physical modelling~\cite{NEURIPS2023_ba748557}. \textit{Chargym} offers an RL-friendly testbed for evaluating cost-driven policies, though it simplifies real-world details by assuming uniform EV types and synthetic demand patterns~\cite{karatzinis2022chargym}.

    To address these gaps, our simulation integrates transformer-aware power modelling, realistic two-stage EV charging curves, degradation-aware battery models, and full V2G support within a Gym-compatible framework~\cite{EV2Gym}. It accommodates a wide variety of control approaches, including heuristic methods, MPC, optimisation, and RL, and incorporates standardised evaluation metrics that enable reproducible and scalable testing under realistic operational constraints.

    \section{Methodology}
    \label{sec:methodology}

    \textcolor{blue}{This section outlines the overall methodological framework adopted in this study. The proposed approach integrates a physics-informed simulation environment, a carbon-aware RL control strategy, and a structured comparative evaluation procedure. Fig.~\ref{fig:system_design} illustrates the integrated system architecture, showing how carbon intensity data, renewable generation profiles, and EV charging models interact under the emission-aware RL algorithm.

    \textbf{Simulation Environment.} All experiments are conducted within EV2Gym~\cite{EV2Gym}, an open-source Gym-compatible platform supporting V2G, transformer-level constraints, realistic two-stage battery charging curves, and degradation-aware battery models. The environment is configured to represent a workplace charging lot with 25 EVSEs fed by a single transformer. EV arrivals, session durations, and energy demands follow empirical workplace profiles from ElaadNL. Time-varying inflexible background loads and a daily demand-response event are included. Full details of the simulator architecture and system modelling are provided in Section~\ref{simulator_modelling}.

    \textbf{Energy and Emissions Modelling.} Carbon emissions are computed at each 15-minute timestep by multiplying the net grid energy draw by the real-time carbon intensity signal sourced from EirGrid~\cite{smartgrid-co2}. Behind-the-meter solar PV and wind generation profiles derived from 2023 Irish data are integrated as zero-carbon offsets to grid demand. Renewable penetration varies from 0\% to 50\% of total daily charging load by scaling installed capacity multipliers. Full formulations are given in Section~\ref{sec:energy_modelling}.

    \textbf{Control Strategies.} Nine control strategies are implemented and evaluated under identical simulation conditions: six heuristic baselines (AFAP, AFAP+, AFAP*, ALAP, FSB, RR), two emission-aware MPC variants (G2V and V2G), and the proposed SAC-based RL agent. The RL agent uses a continuous action space with per-EVSE normalised power setpoints in $[-1, 1]$, an emission-aware state representation including carbon intensity forecasts and renewable generation forecasts over a 20-step horizon, and a multi-objective reward function penalising emissions, unused renewables, and unmet user demand. Full strategy specifications are provided in Section~\ref{charging_strategy}.

    \textbf{Experimental Design.} Five renewable penetration scenarios are evaluated: No RE, 50\% Solar, 50\% Wind, 25\% Hybrid, and 50\% Hybrid. \textcolor{blue}{Throughout this paper, the scenario names Solar, Wind, and Hybrid refer to solar-only, wind-only, and combined solar and wind supply, respectively, with capital letters used consistently in all subsequent references.} Each strategy is run for 10 independent stochastic episodes per scenario, and results are reported as means with standard deviations where applicable. Performance is assessed across five metrics: total CO$_2$ emissions, carbon intensity (gCO$_2$/kWh), transformer overload (kWh), user satisfaction (\%), and renewable self-consumption ratio. This controlled multi-scenario design isolates the effect of the control strategy from that of the renewable supply level.}

    \section{EV2Gym Simulator Architecture and System Modelling}
    \label{simulator_modelling}

    As shown in Fig.~\ref{fig:EV2Gym_Architecture}, EV2Gym operates in three phases: initialisation, where scenarios and parameters are specified; simulation, where EVs, chargers, and transformers interact under Charge Point Operator (CPO) control with user-selected algorithms; and evaluation, where the simulator produces plots, metrics, and replay files. This modular workflow ensures reproducible and extensible benchmarking of smart charging strategies.

    \subsection{EV2Gym Simulator Overview}
    EV2Gym is an open-source simulation platform designed for comprehensive EV smart charging studies with V2G support~\cite{EV2Gym}. It provides a unified Gym-like environment to develop and benchmark control algorithms, including heuristic rules, optimisation approaches such as mixed-integer programming and MPC, and RL, all under realistic operating conditions~\cite{EV2Gym}. The simulator is modular and configurable, allowing users to define custom scenarios or select pre-designed case studies through flexible configuration files. To improve realism, EV2Gym incorporates detailed models of EV batteries, charging stations, and driver behaviour validated with real-world data. This realism extends to CPO considerations such as limited capacity contracts and demand response events, which are often overlooked in simpler tools. By offering a standardised and extensible framework, EV2Gym supports rigorous evaluation of charging strategies at scale while ensuring reproducibility across experiments~\cite{EV2Gym}.
    
    \begin{figure}
        \centering
        \includegraphics[width=0.9\linewidth]{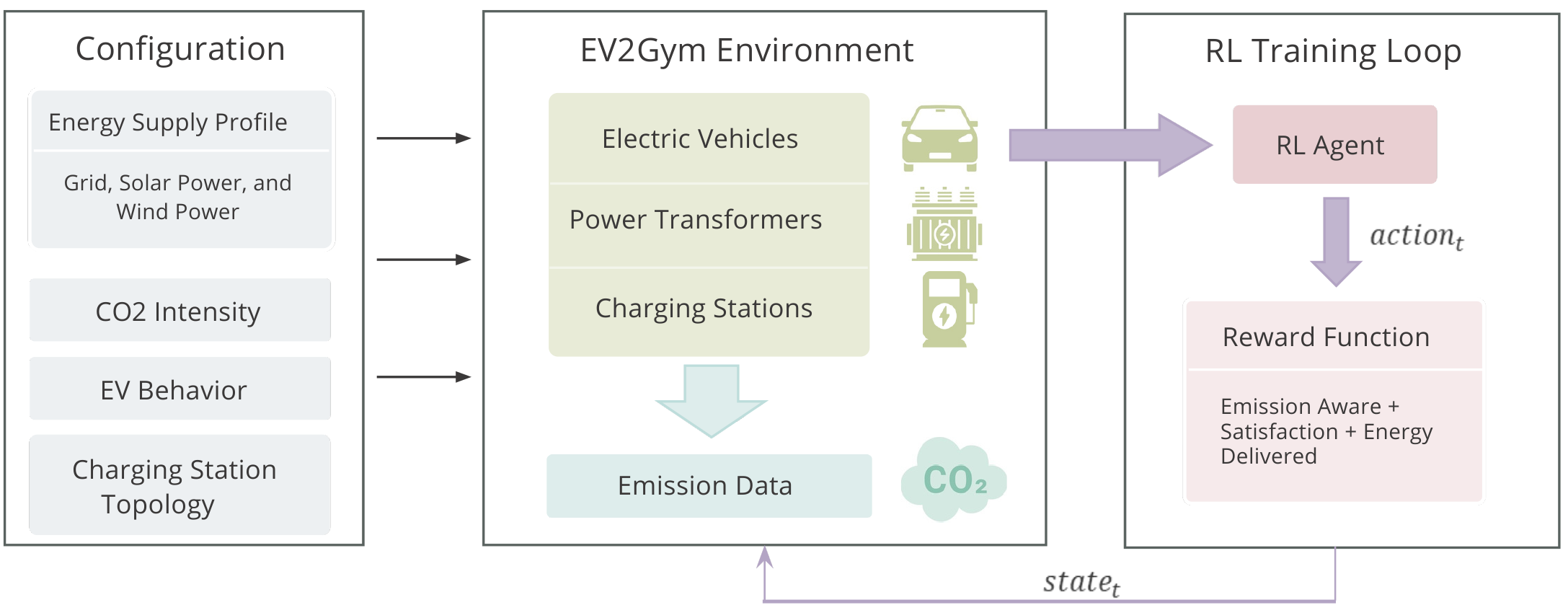}
        \caption{The integrated system architecture incorporates CO$_2$ intensity data, wind generation, and overall energy supply profiles. During simulation, the states of models such as EVs, charging stations, and emissions are updated according to the emission-aware RL algorithm.}
        \label{fig:system_design}
    \end{figure}
    
    \subsection{\textcolor{blue}{Electric Vehicle} and Charger Models}
    In EV2Gym, each EV is modelled with detailed electrical and behavioural parameters that reflect real-world characteristics~\cite{EV2Gym}. Each vehicle has a defined battery capacity and a dynamic state of charge (SoC) that evolves with charging and discharging actions. Charging and discharging are constrained by mode-specific power limits distinguishing AC and DC, and by efficiency factors for both directions, including V2G operations~\cite{EV2Gym}.

    EV arrivals and departures follow stochastic behaviour models. On arrival, each EV is assigned an initial SoC, a departure time, and a target energy level. These parameters are sampled from probability distributions derived from empirical charging data, reproducing realistic patterns in residential, workplace, and public charging contexts~\cite{EV2Gym}.

    Chargers can serve one or more EVs through multiple electric vehicle supply equipment (EVSE) connectors. Each EVSE enforces its own maximum charging and discharging current, while the station is subject to an aggregate current limit across all connectors. This hierarchical constraint structure reflects practical station-level limits such as circuit breaker ratings or transformer thermal thresholds~\cite{10315215}. If a control strategy issues a command beyond these limits, EV2Gym applies clipping or normalisation to maintain feasibility~\cite{EV2Gym}.

    Chargers may be defined as AC or DC, which sets voltage, phase configuration, and bi-directionality. The simulator also supports user-defined charging efficiencies and pricing parameters, enabling consideration of economic factors such as time-of-use tariffs and discharging incentives~\cite{EV2Gym}.

    Together, the EV and charger models provide a high-fidelity representation of charging behaviour and infrastructure constraints. They form a robust foundation for evaluating smart charging algorithms under practical operating conditions.


    \subsection{Transformer and Grid Setup}
    EV2Gym models the local distribution grid through one or more transformers, each delivering power from the grid to a set of connected charging stations~\cite{EV2Gym}. Each transformer has a rated maximum power capacity and may be associated with multiple chargers. At every simulation step, the aggregate charging demand from all EVs connected to a transformer is computed and compared with its capacity limit.

    To reflect real-world infrastructure constraints, the transformer model enforces a hard power cap. If total demand exceeds the rated capacity, the excess is not curtailed but recorded as an overload event~\cite{EV2Gym}. This differs from the per-charger constraint model, where charging rates are automatically adjusted. By recording overloads rather than suppressing them through automatic curtailment, EV2Gym enables post-hoc evaluation of each control strategy's effectiveness in avoiding stressing the distribution network, a critical issue since uncoordinated charging can exceed grid limits~\cite{csengor2018optimal}.

    Beyond EV charging loads, the transformer model includes background inflexible demands and on-site renewable generation such as solar PV. Both factors affect the net available capacity for EV charging and can be configured using real-world datasets or synthetic profiles to represent a range of operating conditions~\cite{EV2Gym}.

    EV2Gym also supports dynamic grid constraints such as demand response (DR) events. Users may define temporary reductions in transformer capacity to mimic grid-side flexibility requests or partial outages. These events can be communicated to control algorithms with minimal lead time to emulate real-time DR signals~\cite{EV2Gym}.

    By combining fixed capacity limits, inflexible loads, renewable variability and time-varying constraints, EV2Gym provides a realistic framework to assess how smart charging algorithms maintain grid stability and mitigate infrastructure stress under evolving system conditions.

    \subsection{Compact Site-Level Mathematical Model}

This subsection summarises the main physical relations governing the site-level charging system. The formulation is adapted from the EV2Gym modelling framework~\cite{EV2Gym}, with notation aligned to the present study and extended to include behind-the-meter wind generation.

Let $c \in \mathcal{C}$ denote a charging station, $p \in \mathcal{P}_c$ an EVSE port, and $t \in \mathcal{T}$ a discrete simulation step of duration $\Delta t$. The electrical power at port $p$ of charging station $c$ is given by~\cite{EV2Gym}
\begin{equation}
    P_{c,p,t} = \eta_{c,p,t}\, I_{c,p,t}\, V_{c,p,t}\sqrt{\phi_{c,p,t}},
    \label{eq:site_evse_power}
\end{equation}
where $I_{c,p,t}$ is the commanded current, $V_{c,p,t}$ is the operating voltage, $\phi_{c,p,t}$ is the number of phases, and $\eta_{c,p,t}$ is the charging or discharging efficiency. Positive values of $P_{c,p,t}$ denote grid-to-vehicle charging, while negative values denote vehicle-to-grid discharging.

Battery charging follows a configurable two-stage model that captures the constant-current (CC) and constant-voltage (CV) regions more realistically than a purely linear approximation~\cite{EV2Gym}. For EV $k$, the state of charge evolves as
\begin{equation}
    \mathrm{SoC}_{k,t} =
    \begin{cases}
        \mathrm{SoC}_{k,t-1} + \dfrac{P_{k,t}\,\Delta t}{\bar{E}_k},
        & \mathrm{SoC}_{k,t-1} < \tau_k, \\[8pt]
        1 + \left(\mathrm{SoC}_{k,t-1}-1\right)
        \exp\!\left(
            \dfrac{P_{k,t}\,\Delta t}{\bar{E}_k(\tau_k-1)}
        \right),
        & \mathrm{SoC}_{k,t-1} \ge \tau_k,
    \end{cases}
    \label{eq:site_two_stage_soc}
\end{equation}
where $\bar{E}_k$ is the battery capacity and $\tau_k \in (0,1]$ is the SoC threshold marking the onset of the CV region. Setting $\tau_k = 1$ reduces the model to a linear approximation~\cite{EV2Gym}.

At the charging-station level, an aggregate current constraint is enforced across all ports of charging station $c$~\cite{EV2Gym}:
\begin{equation}
    I^{\mathrm{cs},-}_{c} \;\le\; \sum_{p \in \mathcal{P}_c} I_{c,p,t} \;\le\; I^{\mathrm{cs},+}_{c},
    \label{eq:site_charger_limit}
\end{equation}
where $I^{\mathrm{cs},-}_{c}$ and $I^{\mathrm{cs},+}_{c}$ are the minimum and maximum allowable aggregate currents, respectively. If the combined EVSE currents exceed these limits, they are normalised to satisfy~(\ref{eq:site_charger_limit})~\cite{EV2Gym}.

The net site power seen by the transformer combines EV demand, inflexible load, and behind-the-meter renewable generation. Adapting the EV2Gym transformer-level balance and extending it to include wind generation, the site-level power balance is written as
\begin{equation}
    P^{\mathrm{site}}_{t}
    =
    \sum_{c \in \mathcal{C}} \sum_{p \in \mathcal{P}_c} P_{c,p,t}
    \;+\;
    L^{\mathrm{inf}}_{t}
    \;-\;
    P^{\mathrm{PV}}_{t}
    \;-\;
    P^{\mathrm{wind}}_{t},
    \label{eq:site_power_balance}
\end{equation}
where $L^{\mathrm{inf}}_{t}$ is the inflexible load, and $P^{\mathrm{PV}}_{t}$ and $P^{\mathrm{wind}}_{t}$ denote the available solar and wind generation, respectively. In this formulation, local renewable generation offsets site demand before any residual is drawn from the grid. Equation~(\ref{eq:site_power_balance}) represents the net site-level power seen by the transformer and is used for infrastructure loading analysis. The EV-specific grid import used later for charging-emissions accounting is defined separately in Section~\ref{sec:behind-the-meter integration}.

Transformer stress is quantified through the overload magnitude
\begin{equation}
    O_t =
    \max\!\left(
        0,\;
        P^{\mathrm{site}}_{t} -
        \left(P^{\mathrm{tr}}_{\max} - P^{\mathrm{DR}}_{t}\right)
    \right),
    \label{eq:site_transformer_overload}
\end{equation}
where $P^{\mathrm{tr}}_{\max}$ is the transformer capacity and $P^{\mathrm{DR}}_{t}$ is any temporary demand-response capacity reduction. A positive value of $O_t$ indicates a capacity violation, which is recorded rather than automatically curtailed~\cite{EV2Gym}.

Together, (\ref{eq:site_evse_power})--(\ref{eq:site_transformer_overload}) provide a compact mathematical description of the charging site, linking battery dynamics, charger constraints, renewable integration, and transformer loading. These relations form the physical basis for the emissions model and control formulations developed in the following sections.

    \subsection{Simulation Configuration}
    All experiments in this study follow a standardised setup to ensure fair and consistent comparison across charging strategies. The EV2Gym environment operates in discrete time steps of fixed duration, typically 15 minutes, which balances temporal resolution and computational efficiency~\cite{EV2Gym}. Each simulation spans 24 hours, yielding 96 steps that capture the full diurnal cycle of EV activity and grid dynamics. Initialisation allows users to specify the starting date and time, enabling alignment with specific day types or seasonal variations. This supports the use of representative arrival and departure profiles as well as renewable generation patterns, such as solar PV output, that vary over time~\cite{EV2Gym}.  

    \textcolor{blue}{Key parameters, including the number of chargers and EVSE connectors, transformer capacity, background load profiles and EV behaviour scenario, are defined before runtime and remain fixed across experiments unless explicitly varied. Table~\ref{tab:ev2gym_params} summarises the main EV2Gym configuration parameters used in this study, including EV battery assumptions, SoC settings, charger ratings, efficiencies, and charger type. This ensures that performance differences arise solely from the control strategies under evaluation.} At each step, the control agent (heuristic, MPC, or RL) generates charging or discharging actions for connected EVs, as shown in Fig.~\ref{fig:system_design}. The environment then updates internal states, including vehicle SoCs, grid load levels, and transformer energy flows, in response to these actions~\cite{EV2Gym}.  

    At the end of each run, EV2Gym computes performance metrics such as total energy delivered, peak transformer load, overload events, and user satisfaction. It can also produce replay files and graphical outputs for further analysis~\cite{EV2Gym}. By maintaining uniform configuration parameters across all strategies, the framework enables rigorous and reproducible benchmarking of smart charging methods under realistic operating conditions.

        \begin{table}[t]
    \centering
    \caption{EV2Gym simulation parameters used in this study.}
    \label{tab:ev2gym_params}
    \begin{tabular}{p{0.45\linewidth} p{0.47\linewidth}}
        \toprule
        \textbf{Parameter} & \textbf{Value} \\
        \midrule
        Battery capacity range              & Heterogeneous: based on EV models~\cite{EV2Gym}; Default: 10-50 kWh        \\
        Initial SoC range                   & Stochastic (ElaadNL workplace data~\cite{EV2Gym})       \\
        Target SoC                          & 85\%                                \\
        SoC bounds (min\,/\,max)            & 0\%\,/\,100\%                       \\
        CC/CV threshold $\tau$              & 1.0 (linear model)                   \\
        \midrule
        Max charging power per EVSE         & 22 kW   \\
        Max discharging power per EVSE      & 22 kW   \\
        Charging efficiency $\eta$          & 100\%                                \\
        Discharging efficiency $\eta$       & 100\%                                \\
        Charger type                        & AC (400\,V, 3-phase)                 \\
        \bottomrule
    \end{tabular}
\end{table}

    \section{Energy and Emissions Modelling}
    \label{sec:energy_modelling}
    Accurately modelling energy flows and associated emissions is essential for evaluating the sustainability of EV charging strategies. This section describes how EV2Gym represents carbon intensity signals, renewable generation profiles, and varying penetration levels to quantify the environmental impact of different charging scenarios.
    \subsection{Carbon Emissions Model}
    Accurately representing emissions is critical for assessing the sustainability of EV charging strategies. This work incorporates a carbon emissions model that links charging activity to the time-varying carbon intensity of the grid. This enables the evaluation of how different scheduling decisions influence overall emissions and supports the development of carbon-aware control strategies.
    \subsubsection{Dynamic Grid Intensity}
    Real-time carbon intensity data from EirGrid for 2023 is used to represent the emissions factor of grid electricity at each timestep. The original data is reported in grams of CO$_{2}$ per kilowatt-hour (gCO$_{2}$/kWh) but is converted into kilograms of CO$_{2}$ per kilowatt-hour (kgCO$_{2}$/kWh) for consistency in subsequent calculations. The intensity value updates every 15 minutes, reflecting changes in the generation mix, such as the share of renewables versus fossil fuels~\cite{smartgrid-co2}. By treating carbon intensity as a time-dependent input, the simulation can determine whether charging occurs during high-carbon or low-carbon periods and quantify the resulting emissions.

    \subsubsection{Per-Step Emission Calculation}
    At each simulation step $t$, the model computes emissions by multiplying the grid CO$_{2}$ intensity by the net energy drawn from the grid during that interval. \textcolor{blue}{This approach follows the general emissions accounting framework established by the IPCC, where emissions are calculated as the product of activity data and an emission factor~\cite{ipcc2006guidelines}, and is consistent with the GHG Protocol Scope 2 Guidance for quantifying indirect emissions from purchased electricity~\cite{ghgprotocol2015scope2}.}  Formally, this is expressed as:
    \begin{equation}
        E_{\mathrm{CO}_2,t}= I_{t}\times E_{\mathrm{grid},t}\label{eq:co2_emissions}
    \end{equation}

    where $I_{t}$ (kg CO$_2$/kWh) represents the grid carbon intensity at time $t$, and $E_{\mathrm{grid},t}$ (kWh) is the electrical energy supplied by the grid in that interval. If carbon intensity data is reported in grams CO$_2$/kWh, it is converted to kilograms by dividing by 1000. This calculation yields the kilograms of CO$_2$ emitted in that step.

    This per-step formulation reflects that charging-related emissions depend on when an EV charges, because grid carbon intensity varies over time~\cite{smartgrid-co2}. Charging during periods of high carbon intensity produces more emissions than charging when the grid mix is cleaner. By incorporating a time-resolved carbon intensity signal, the model quantifies the environmental impact of each charging decision, providing the foundation for the RL reward function to encourage alignment of charging with low-emission grid periods. Such carbon-aware scheduling is therefore important for reducing charging-related emissions by aligning EV charging with lower-carbon operating periods~\cite{NationalEVAnnualReport}.

    \subsection{Renewable Generation Profiles}
    To capture the influence of local renewables on EV charging, this work models behind-the-meter solar and wind generation. These resources offset grid demand at each timestep, allowing analysis of how varying renewable penetration affects energy usage and emissions.
    \subsubsection{Solar \textcolor{blue}{Photovoltaic} Generation}
    Solar PV is integrated as a behind-the-meter resource that directly offsets grid demand for EV charging~\cite{EV2Gym}. The generation profile is based on 2023 Irish solar data from the ENTSO-E Transparency Platform~\cite{pvgis-solar}, which provides high-resolution statistics on generation and demand across Europe. Profiles capture daily and seasonal variation: summer days exhibit strong midday generation, while winter evenings show little or none.  

    In the simulation, the solar profile is scaled by a configurable capacity multiplier to represent different levels of on-site PV installation. The resulting generation is subtracted from the charging load at each timestep, and any surplus beyond immediate charging needs is curtailed, consistent with a local behind-the-meter setup. By using real time-series profiles, the model preserves key features such as evening residual load (low solar availability) and periods when solar output does not align with charging demand. This realism is essential because integrating variable renewables remains challenging due to intermittency and forecast uncertainty~\cite{9246271}.

    \subsubsection{Wind Generation}
    Wind is incorporated as an additional behind-the-meter source to offset EV charging demand. Profiles for 2023 are derived from the EirGrid Smart Grid Dashboard~\cite{smartgrid-co2}, which reports historical and real-time wind output. Irish wind farms exhibit large swings in output, as many sites are geographically clustered and experience similar wind conditions simultaneously~\cite{1709436}. This co-location effect results in ramps and drops in generation over short intervals, which must be considered in charging strategies.  

    As with solar PV, wind generation is subtracted from the charging load at each timestep, and any surplus beyond immediate needs is curtailed. By combining solar and wind profiles, the model analyses how varying renewable penetration levels affect grid energy use and emissions.

    \subsubsection{Behind-the-Meter Integration}
    \label{sec:behind-the-meter integration}
    Both solar and wind are modelled as behind-the-meter resources that directly serve EV charging. \textcolor{blue}{This behind-the-meter integration follows the same approach used in the EV2Gym simulator~\cite{EV2Gym}, where on-site renewable generation is netted against charging demand within the local power balance before any grid interaction is considered. At each timestep, the available renewable power is first allocated against EV charging demand, so that the remaining EV-related grid import is}
        \begin{equation}
        P^{\mathrm{EV,grid}}_t
        =
        \max\!\left(
        0,\,
        P^{\mathrm{EV,ch}}_t-(P^{\mathrm{PV}}_t+P^{\mathrm{wind}}_t)
        \right),
        \label{eq:ev_grid_import}
        \end{equation}

    The max function ensures that any excess renewable generation is curtailed. This reflects site-level operation, where surplus renewable energy is not exported, and mirrors system-level practice during oversupply when operators curtail renewables~\cite{10253224}.

    By modelling integration in this way, combined solar and wind directly reduce grid electricity use and emissions. For example, if fleet demand is 50 kW at noon and solar and wind generate 30 kW and 10 kW respectively, the grid supplies only 10 kW while 40 kW is met locally. Over a typical day, solar contributes during daylight, while wind often provides power in the evenings or overnight. This combined supply reduces reliance on the grid and lowers total CO$_2$ emissions.

    \subsection{Renewable Penetration Levels}
    We simulate renewable penetration levels ranging from 0\% to 50\% of the total EV charging load to assess how increasing on-site renewable capacity affects grid usage and associated emissions. Penetration is controlled by scaling installed solar PV and wind capacities relative to baseline charging demand. In EV2Gym, this is specified through configuration parameters~\cite{EV2Gym}. A multiplier of 1.0 represents the nominal solar capacity, while higher values proportionally increase generation. For example, a value of 2.0 doubles PV output relative to the base case.

    Wind generation is scaled analogously, enabling flexible combinations of solar and wind penetration. A 0\% renewable case sets both multipliers to zero, meaning all charging energy is supplied by the grid. In contrast, a 100\% case uses sufficiently high multipliers so that during peak generation, renewables can fully cover charging demand, minimising the need for grid power.

    Intermediate penetration scenarios (e.g., 25\% Hybrid, 50\% Solar, or 50\% Wind) are created by selecting multipliers that achieve a target share of daily charging energy from renewables. The offset is computed dynamically at each timestep: higher renewable penetration reduces grid consumption and lowers emissions, since behind-the-meter generation is treated as zero-carbon.

    Tracking emissions across scenarios illustrates the environmental benefits of on-site renewables. Under high penetration, midday charging may be fully supplied by local generation with zero grid emissions, whereas in a zero-renewables scenario, the same demand incurs full emissions at the prevailing grid intensity~\cite{EV2Gym}. This modelling approach quantifies how increased renewable integration progressively drives charging-related emissions toward zero. Examples include deploying rooftop solar, on-site wind, or hybrid systems.

    Emissions are calculated at each timestep and aggregated over the simulation horizon, yielding total CO$_2$ emissions for each scenario as a key outcome metric. This framework provides a transparent basis to evaluate decarbonisation strategies for EV charging, including shifting charging to lower-carbon grid periods and expanding local renewable capacity.

    \section{Charging Strategy and Scenario Design}
    \label{charging_strategy}
    \subsection{Baseline Control Strategies}
    To evaluate the performance of advanced optimisation and learning-based methods, we establish baseline charging strategies that range from simple heuristics to MPC.
    \subsubsection{Heuristic Methods}
    We first consider simple scheduling rules that serve as baseline strategies for comparison. The most basic is Charge As Fast As Possible (AFAP), which charges each EV at maximum power upon arrival without regard to network constraints or grid conditions. AFAP represents an uncontrolled “greedy” policy and is widely used in the literature as a reference for worst-case grid stress and emissions when no coordination is applied~\cite{EV2Gym}.  
    
    Three variants of AFAP are also implemented:  
    \begin{itemize}
        \item AFAP with Power Limit (AFAP+): AFAP with an additional power cap at each timestep, ensuring that aggregate charging demand does not exceed a specified threshold~\cite{EV2Gym}.  
        \item AFAP to Desired Level (AFAP*): Charging proceeds at maximum rate only until a specified state of charge or energy level is reached, after which charging stops.  
        \item Fixed Scheduled Baseline (FSB): Charging begins at a predetermined start time, allowing alignment with grid conditions, time-of-use pricing, or renewable availability~\cite{GANESH2022111833}.  
    \end{itemize}  

    We also consider Charge As Late As Possible (ALAP), which delays charging until the latest feasible start time that still guarantees the required energy level by departure. This rule has been studied as a way to shift demand away from peak periods when grid conditions can be forecast~\cite{GANESH2022111833}.  

    Finally, a Round Robin (RR) policy allocates available capacity in rotation among connected EVs. RR ensures fairness while capping the total site load to the station’s power limit at each timestep. Strategies such as RR and ALAP are easy to implement and can modestly improve load balancing relative to AFAP, but they do not explicitly optimise for emissions or renewable utilisation~\cite{EV2Gym}.  

    Together, these heuristics form the foundation for baseline comparisons. They illustrate how incremental constraints or scheduling rules can reduce peak demand and grid stress, while establishing benchmarks against which more advanced optimisation and RL methods can be evaluated.

    \subsubsection{Model Predictive Control}
    In this work, the baseline MPC controllers in EV2Gym were extended with an emission-aware objective: one for unidirectional grid-to-vehicle (G2V) operation and another for bidirectional V2G. Each solves a mixed-integer optimisation problem over a receding horizon $H$. At each timestep, the controller observes the current state of all EVs (SoC, arrival/departure) and external forecasts (inflexible load, PV generation, and grid CO$_2$ intensity), formulates an MILP, and solves it with the Gurobi optimiser~\cite{EV2Gym}. Only the first-step control actions (charging or discharging power for each EV) are applied, and the process repeats in a moving-horizon fashion.  

    The MPC formulation enforces standard EV and grid constraints, following the EV2Gym baseline~\cite{EV2Gym}. Each EV must satisfy battery constraints (SoC bounds and departure requirements) as well as per-EV charging and discharging limits. Station-level charger limits and transformer capacity constraints ensure that aggregate EV demand plus inflexible load never exceeds rated limits. Binary decision variables prohibit simultaneous charging and discharging at the same EVSE. In short, the constraint structure and system dynamics follow the EV2Gym baseline~\cite{EV2Gym}.  

    The objectives differ by mode. In G2V, the cost function minimises the squared tracking error relative to a target setpoint. In V2G, it maximises CPO profit (charging and discharging revenues) while satisfying user demand. Both are extended with an emission-aware term: at each horizon step, the cost is weighted by the forecasted grid CO$_2$ intensity multiplied by net energy exchange. This biases scheduling towards periods of low carbon intensity and reduces emissions. This emission-aware formulation is a novel extension beyond the default EV2Gym MPC.

    \subsection{\textcolor{blue}{Reinforcement Learning} Environment and Reward Design}

    \begin{figure}     
    \centering
    \includegraphics[width=0.4\textwidth]{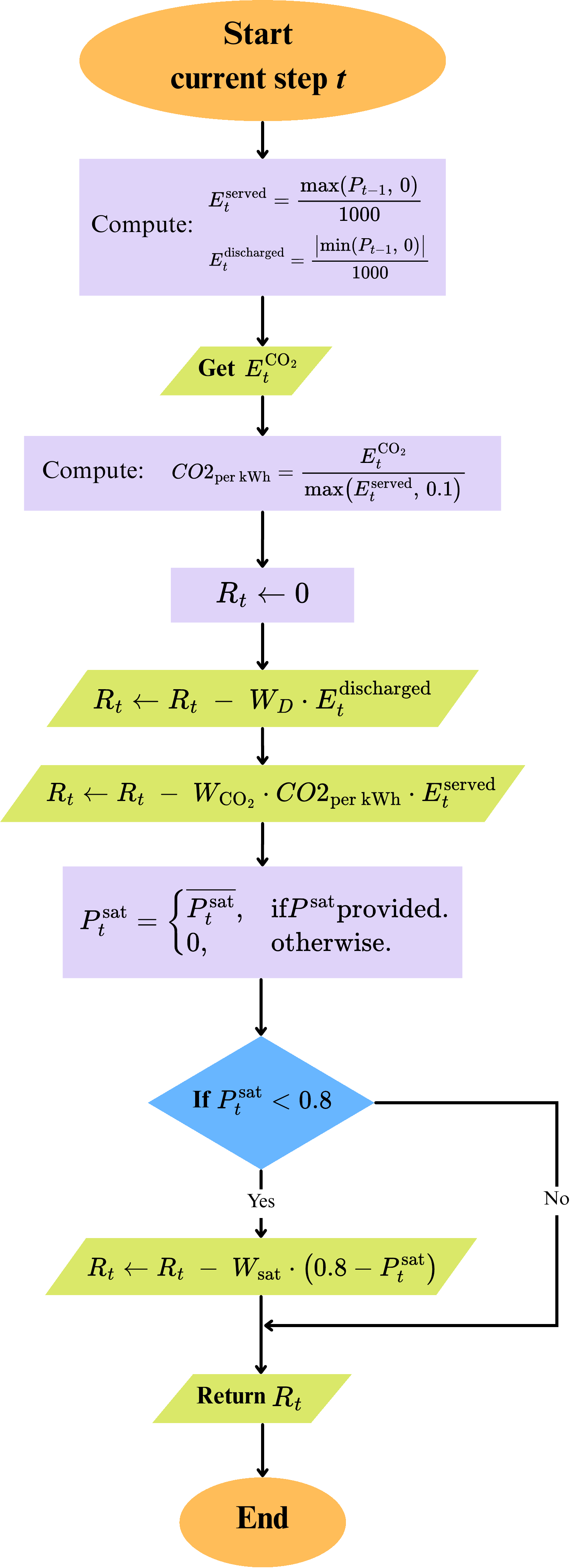}%
    \caption{Reward function flowchart.}
    \label{fig:flow_chart}    
    \end{figure}

    Here, we employ the Soft Actor-Critic (SAC) algorithm from the Stable-Baselines3 library~\cite{raffin2021stable}, extended with customised reward and state functions.

    \subsubsection{Emission-Aware State Function Design}
    In this work, we propose an emission-aware state representation for smart EV charging control. The state at timestep $t$ is defined as:

    \begin{equation}
        \begin{aligned}
            \mathbf{S}_{t}={} & \left[ t,\; P^{\mathrm{tot}}_{t-1}\right] \cup \left[ \forall w:\ \big( L^{\mathrm{inf}}_{w,t}~,\ P^{\mathrm{RE}}_{w,t}~,\  \tilde L^{\mathrm{inf}}_{w,~t:t+H}~,\ \tilde P^{\mathrm{RE}}_{w,~t:t+H}\big) \right] \\
                              & \cup \left[ \tilde{E}^{\mathrm{CO}_2}_{t:t+H}\right] \cup \left[ \forall c \in C,\ \forall p \in P_{c}:\ \big( \mathrm{SoC}_{cp}~,\ \mathrm{ToD}_{cp}\big) \right]
        \end{aligned}
    \end{equation}

    \begin{algorithm}
    \small
        \caption{Pseudo Code for State Function}
        \label{alg:state-concise}
        \begin{algorithmic}
            [1] \Require $H{=}20$;\; env with \Statex\hspace{\algorithmicindent}$\{
            t,\;P^{\mathrm{tot}}_{t-1},\;\texttt{transformers},\;\texttt{charging\_stations
            },\;\tilde E^{\mathrm{CO}_2}\}$ \State $t \gets \min(\texttt{current\_step
            },\,\texttt{simulation\_length}{-}1)$ \State $S_{t}\gets [\,t,\,P^{\mathrm{tot}}
            _{t-1}(t)\,]$ \For{$w \in \texttt{transformers}$} \State
            $S_{t}\mathrel{+\!=}\textsc{Scale100}\big(L^{\mathrm{inf}}_{w,t},\,P^{PV}
            _{w,t},\,P^{Wind}_{w,t}\big)$
            \State
            $(\tilde L,\,\tilde P^{PV}) \gets \texttt{get\_load\_pv\_forecast}(w,
            \,t,\,H)$
            \State
            $(\cdot,\,\tilde P^{Wind}) \gets \texttt{get\_load\_wind\_forecast}(w
            ,\,t,\,H)$
            \State $S_{t}\mathrel{+\!=}\textsc{Scale100}\big($ \Statex\hspace{\algorithmicindent}$\textsc
            {Pad}(\tilde L,\,H),\ \textsc{Pad}(\tilde P^{PV},\,H),$
            \Statex\hspace{\algorithmicindent}$\textsc{Pad}(\tilde P^{Wind},\,H)\big
            )$
            \EndFor \State $S_{t}\mathrel{+\!=}\textsc{Scale1000}\big($ \Statex\hspace{\algorithmicindent}$\textsc
            {Pad}(\tilde E^{\mathrm{CO}_2}_{\,t:t+H-1},\,H)\big)$
            \For{$c \in \texttt{charging\_stations}$} \For{$p \in P_{c}$} \State
            $(SoC_{cp},\,ToD_{cp}) \gets$ \Statex\hspace{\algorithmicindent}$\begin{cases}
                (p.\texttt{get\_soc}(),\ \max(p.\texttt{time\_of\_departure}{-}t,\,0)), & p\neq\texttt{None} \\
                (0,0),                                                                  & \text{otherwise}
            \end{cases}$
            \State $S_{t}\mathrel{+\!=}(SoC_{cp},\,ToD_{cp})$ \EndFor \EndFor
            \State \Return $\texttt{as\_array}(S_{t})$ \Statex
        \end{algorithmic}
    \end{algorithm}

    As shown in Algorithm~\ref{alg:state-concise}, the state function is designed to provide the RL agent with comprehensive information to minimise CO$_2$ emissions while meeting charging requirements. Table~\ref{tab:statevector} summarises the state components, which include normalised time, grid and renewable forecasts, and detailed EV-specific information. The temporal features consist of the current discrete timestep $t$ and the last-period total power $P^{tot}_{t-1}$, anchoring the agent’s decisions in both episode progression and recent aggregate behaviour. The state also incorporates the current inflexible load and the combined output of solar PV and wind generation, providing a real-time view of non-EV demand and available renewable supply at each transformer.  

    To support anticipation of future conditions, the agent receives forecasts of inflexible loads and renewable generation over a lookahead horizon $H$, enabling proactive scheduling during periods of surplus renewable energy or low background demand. In addition, the state includes forecasts of grid emission factors (kg CO$_2$/kWh) over the same horizon, offering foresight into cleaner or more carbon-intensive periods and allowing alignment of charging with low-emission windows. Finally, each EV’s SoC and ToD are included to ensure that user requirements are respected. By knowing how much energy each EV requires and how long it will remain connected, the agent can balance emissions reduction objectives with the timely fulfilment of user charging needs.  

    This structure provides the operational, predictive, and environmental context necessary for the RL policy to coordinate large-scale EV charging in a way that actively reduces carbon emissions while maintaining service quality for drivers.

    \begin{table}
        \centering
        \caption{State vector components for emission-aware smart charging.}
        \label{tab:statevector}
        \begin{tabular}{p{0.18\linewidth} p{0.74\linewidth}}
            \toprule \textbf{Variable}           & \textbf{Description}                                        \\
            \midrule $t$                         & Current discrete time step                                  \\
            $P^{\mathrm{tot}}_{t-1}$             & Last-period total power usage (kW)                          \\
            $L^{\mathrm{inf}}_{w,t}$             & Current inflexible load at transformer $w$ (scaled)         \\
            $P^{\mathrm{RE}}_{w,t}$              & Current renewable generation at transformer $w$ (scaled)    \\
            $\tilde{L}^{\mathrm{inf}}_{w,t:t+H}$ & Forecasted inflexible loads for horizon $H$                 \\
            $\tilde{P}^{\mathrm{RE}}_{w,t:t+H}$  & Forecasted renewable generation for horizon $H$             \\
            $\tilde{E}^{CO_2}_{t:t+H}$           & Forecasted grid CO$_{2}$ emission factors                   \\
            $SoC_{cp}$                           & State of charge of EV at port $p$ in charging station $c$   \\
            $ToD_{cp}$                           & Time to departure (steps) for EV at port $p$ in station $c$ \\
            \bottomrule
        \end{tabular}
    \end{table}

    \subsubsection{\textcolor{blue}{Action Space Design}}
    \textcolor{blue}{At each timestep $t$, the SAC agent outputs a continuous action vector $\mathbf{A}_{t}$ comprising one charging command per active EVSE port:}
    \begin{equation}
        \mathbf{A}_{t} = \left[ a_{cp} \right]_{\forall c \in C,\, \forall p \in P_{c}}, \quad a_{cp} \in [-1,\; 1]
        \label{eq:action_space}
    \end{equation}
    \textcolor{blue}{where $a_{cp}$ is the normalised power setpoint for port $p$ of charging station $c$. Positive values correspond to grid-to-vehicle (G2V) charging, and negative values correspond to vehicle-to-grid (V2G) discharging. The raw action is mapped to physical power by multiplying by the maximum charging or discharging power of the EVSE, and EV2Gym clips any resulting command that would violate per-port or station-level current limits~\cite{EV2Gym}. An action of zero implies the port is idle. This continuous, bounded representation is well-suited to SAC, whose policy parameterises a Gaussian distribution over $[-1,1]^{|P|}$, where $|P|$ is the total number of active EVSE ports. The approach follows the continuous action formulation adopted in recent RL-based EV charging studies~\cite{rossi2025smart,QIU2023113052}, and extends it to the bidirectional (V2G-enabled) setting modelled here.}

    \subsubsection{Reward Function Design}
    The reward function \textcolor{blue}{proposed in this work builds on the penalty-based treatment established in prior safe RL and EV charging formulations.} It is carefully crafted to balance multiple objectives in emission-aware smart charging, including maximising solar energy usage, minimising grid emissions, and ensuring EV charging satisfaction. At each time step $t$, the immediate reward $R_{t}$ is defined as a weighted sum of these components, as shown below:

    \begin{equation}
        \begin{aligned}
            R_{t}={} & - W_{D}\,E^{\text{discharged}}_{t}- W_{\mathrm{CO}_2}\left( \frac{E^{\mathrm{CO}_2}_{t}}{\max\!\left(E^{\mathrm{served}}_{t},\,0.1\right)}\right) \\
                     & \cdot E^{\mathrm{served}}_{t}- P^{\mathrm{sat}}_{t}.
        \end{aligned}
    \end{equation}

    This formulation comprises three penalty terms, each targeting a specific objective, as Table~\ref{tab:reward_vars} details below.

    \begin{table}
        \centering
        \caption{Reward function variables and parameters.}
        \label{tab:reward_vars}
        \begin{tabular}{p{0.20\linewidth} p{0.72\linewidth}}
            \toprule \textbf{Symbol}               & \textbf{Description}                                          \\
            \midrule $E^{\mathrm{discharged}}_{t}$ & Total energy discharged to grid (kWh)                         \\
            $E^{\mathrm{served}}_{t}$              & Total energy served to EVs (kWh)                              \\
            $E^{\mathrm{CO}_2}_{t}$                & CO$_{2}$ emissions from grid energy at $t$                    \\
            $P^{\mathrm{sat}}_{t}$                 & Penalty applied if user satisfaction $< 0.8$                  \\
            $S_{t}$                                & Average user satisfaction at time $t$                         \\
            $W_{D}$                                & Discharging penalty weight (hyperparameter)                   \\
            $W_{\mathrm{CO}_2}$                    & CO$_{2}$ emissions penalty weight (hyperparameter)            \\
            $W_{\mathrm{sat}}$                     & User satisfaction penalty weight (via $P^{\mathrm{sat}}_{t}$) \\
            \bottomrule
        \end{tabular}
    \end{table}

    As shown in Fig.~\ref{fig:flow_chart}, the first term represents the energy (kWh) discharged or curtailed at time $t$, such as excess solar generation not used for charging. A large weight $W_{D}$ is applied to strongly penalise this wasted renewable energy. For example, with $W_{D}=50$, each unused kilowatt-hour incurs a $-50$ reward penalty. This negative incentive encourages the agent to maximise on-site PV self-consumption by charging vehicles with surplus solar power rather than allowing it to be curtailed or exported.

    The second term penalises CO$_{2}$ emissions associated with charging. Let $E^{\text{served}}_{t}$ denote the total energy delivered to EVs at time $t$, and $E^{\text{CO}_2}_{t}$ the corresponding emissions. The ratio $\frac{E^{\text{CO}_2}_{t}}{\max(E^{\text{served}}_{t},0.1)}$ captures the effective carbon intensity (kgCO$_2$/kWh) of the electricity used, and multiplying by $E^{\text{served}}_{t}$ yields the total emissions. A scaling factor $W_{\text{CO}_2}$ adjusts this penalty. For instance, if $W_{\text{CO}_2}=5$ and 1 kWh of charging results in 0.7 kg of CO$_2$, the reward is reduced by approximately $5 \times 0.7 = 3.5$. This term biases charging decisions toward periods of lower carbon intensity and reduces reliance on carbon-intensive supply.

    To ensure user requirements are met, a satisfaction penalty $P^{\text{sat}}_{t}$ is included whenever the average charging satisfaction falls below a defined threshold, as formalised in Equation~(\ref{eq:satisfaction_penalty}). Satisfaction is measured as the fraction of each EV’s demand met by its departure time. If average satisfaction at time $t$ drops below the threshold (e.g., 0.8), the penalty is proportional to the shortfall. For example, with a threshold of 0.8 and a penalty factor of 50, an average satisfaction of 0.7 (10\% below target) results in a $-50 \times (0.8 - 0.7) = -5$ penalty. This threshold-based approach simplifies the reward function while safeguarding the primary objective of ensuring EV availability~\cite{10688856}. Although more advanced safe RL methods could address this via constrained MDP formulations, the penalty-based design remains effective and straightforward to implement.

    \begin{equation}
        P^{\mathrm{sat}}_{t}=
        \begin{cases}
            W_{\mathrm{sat}}\cdot (0.8 - S_{t}), & \text{if }S_{t}< 0.8, \\[4pt]
            0,                                   & \text{otherwise}.
        \end{cases}
        \label{eq:satisfaction_penalty}
    \end{equation}

    Each weight $W_{D}$, $W_{\text{CO}_2}$, and $W_{\text{sat}}$ is chosen based on experimental results to reflect the relative importance of the three objectives: discouraging unnecessary V2G discharging, minimising carbon emissions, and preserving user trust by guaranteeing charging requirements. A large discharging weight $W_{D}$  strongly discourages V2G actions, reflecting the practical cost of battery cycling and the risk of offsetting the emission benefits of charging if the energy discharged is not used productively. The emission penalty weight $W_{\text{CO}_2}$ provides a consistent environmental signal that nudges the agent to prefer charging during low-carbon grid periods or when on-site solar is available. For example, with $W_{\text{CO}_2}$ set to 5, charging 10~kWh when the grid carbon intensity is 0.5kg~CO$_2$/kWh results in an emission penalty of about 25 reward points, which is significant but not so large as to dominate the overall reward. Finally, setting $W_{\text{sat}}$ equal to $W_{D}$ keeps the satisfaction penalty on the same scale as the discharging penalty, ensuring that the agent cannot gain trivial emission reductions or penalty avoidance by sacrificing user charging needs. These specific values were determined after evaluating multiple experimental configurations and selecting the ones that yielded the most balanced trade-offs.

    Our reward function's multi-term structure is specifically designed to balance environmental objectives with operational constraints. In contrast to a reward function focused solely on emissions, which penalises CO$_{2}$ but may overlook renewable utilisation, or a charging-speed strategy such as AFAP, which maximises user satisfaction through immediate full-power charging while ignoring grid impact, our formulation provides a more holistic incentive. By explicitly penalising unused solar energy, the agent is directly encouraged to utilise available PV generation rather than drawing from the grid or curtailing renewables. Simultaneously, the CO$_{2}$ penalty steers the agent toward low-carbon energy usage, and the satisfaction term safeguards the quality of service. This carefully weighted combination makes our reward function uniquely effective in guiding the learning agent to achieve high PV self-consumption and low emissions without compromising EV charging requirements.

    \subsection{Simulation Scenarios}
    All simulations use a common EV2Gym configuration representing a workplace charging lot with 25 EVSEs (single transformer) and heterogeneous EVs. Vehicle arrivals, time-of-stay, and energy needs follow the realistic workplace profile from ElaadNL's open data. V2G functionality is enabled, and inflexible loads and a single per-day demand-response event are included in the transformer model~\cite{EV2Gym}. In other words, the charger count, transformer limits, EV behaviour, and demand-response schedule are identical across all cases, ensuring a controlled base scenario.

    We then define three main scenarios, differing only in the on-site renewable generation assumptions:
    \begin{itemize}
        \item Baseline: No renewable generation. Both PV and wind are disabled in the configuration, so the only supplies are EV charging and inflexible load. This essentially reflects uncontrolled charging (e.g. the AFAP heuristic) under a pure grid-dominated scenario~\cite{EV2Gym}.

        \item Smart Charging: Same as Baseline (no PV or wind) but using a smart charging control algorithm. That is, we apply advanced scheduling to the chargers while keeping the grid assumptions identical to Baseline. This isolates the effect of the smart charging strategy itself without any renewable contribution.

        \item Smart Charging + Renewables: Same control algorithm as in case (2), but with local renewables enabled. We consider three renewable variants corresponding to Solar (PV only), Wind (wind only), and Hybrid (combined PV and wind) configurations. In each variant, we use realistic solar and wind profiles.
    \end{itemize}

    The first scenario establishes a baseline for uncontrolled charging, while the second demonstrates the impact of smart scheduling without renewables. The third scenario introduces local renewable generation, allowing us to assess how on-site solar and wind resources can further reduce grid energy usage and carbon emissions when combined with advanced control strategies.

    Each scenario is run for a fixed period, 24 hours (starting from 5am), to capture daily patterns in EV charging behaviour, grid conditions, and renewable generation. The results are then compared across scenarios to quantify the benefits of smart charging and renewable integration in terms of total CO$_{2}$ emissions, grid energy usage, and user satisfaction.

    We evaluate all charging strategies using a common set of performance metrics, enabling fair comparison across the Baseline, Smart Charging, and Smart+Renewables scenarios. The metrics are as follows:

    \begin{itemize}
        \item \textbf{Total CO$_{2}$ Emissions (kg):} The total carbon emissions from EV charging, computed by summing the product of net grid energy draw and the time-varying grid CO$_{2}$ intensity over all timesteps. At each timestep, the net grid draw (charging minus any V2G export) is multiplied by the CO$_{2}$ intensity forecast as shown in Equation~\ref{eq:co2_emissions}, and these values are summed to yield total emissions. This is a user-defined metric (not built into EV2Gym) that captures the carbon impact of the charging strategy.

        \item \textbf{Grid Energy Usage (kWh): } The total electrical energy drawn from the grid for EV charging, ignoring any V2G exports. In practice, this is the sum of all positive EV charging power over time (i.e. the net energy delivered to vehicles from the grid). In EV2Gym this corresponds to the “Total Energy Charged” metric $E_{ch}$(the time-integral of charging power $P_{ch}$)~\cite{EV2Gym}. We report this in kWh. 
        
        \item \textbf{Transformer Overload Count:} The number of timesteps during which the power draw from any transformer exceeded its rated capacity. EV2Gym tracks transformer power and flags overloads whenever the instantaneous draw exceeds the transformer's limit~\cite{EV2Gym}. We derive this metric by counting all timesteps with at least one overload event.
        
        \item \textbf{User Satisfaction(\%):} The average, over all EVs, of the ratio of actual SoC at departure to the desired (target) SoC, expressed as a percentage. This follows EV2Gym's definition of $\epsilon_{\text{usr}}$: \begin{equation} \epsilon_{\text{usr}}= \frac{1}{|E|}\sum_{k \in E}\frac{\text{SoC}_{k}}{\text{SoC}^{*}_{k}} \label{eq:epsilon_usr} \end{equation} where $\text{SoC}_{k}$ is the energy delivered to EV $k$ by departure and $\text{SoC}^{*}_{k}$ is its desired energy. (We multiply by 100\% to express this ratio as a percentage.) A score of 100\% means all users reach their desired charge. This metric is provided by EV2Gym~\cite{EV2Gym}.

        \item \textbf{RE Self-Consumption Ratio (\%)} The fraction of local renewable generation (PV and/or wind) that is directly consumed by EV charging (i.e. not exported to the grid or curtailed). We compute this by summing all EV charging power that comes from on-site renewable sources and dividing by the total local generation, then multiplying by 100\%. It quantifies how effectively the charging strategy utilises available renewables.
    \end{itemize}

    All of the above metrics are either reported directly by EV2Gym (e.g. $\epsilon_{\text{usr}}$) or computed from the simulation logs of EV charging and transformer power. By using this unified metric set in all experiments, we ensure a consistent basis for comparing strategy performance under varying renewable availability and load conditions~\cite{EV2Gym}. This allows us to directly assess trade-offs (e.g. between emissions reduction and user satisfaction) across the Baseline, Smart Charging, and Smart+Renewables cases.

    \section{Results and Discussion}
    \label{results}
    This section evaluates the charging strategies under multiple scenarios and discusses their performance with respect to user needs and grid impact. We compare (i) simple baseline charging heuristics, (ii) an MPC strategy, and (iii) the proposed RL strategy. Key performance metrics include user satisfaction, CO$_2$ emissions, total energy charged/discharged, transformer overload, and RE ratio. Figures and tables are referenced with generic labels (e.g., Fig.~\ref{fig:EV_energy_level}, Table~\ref{tab:rl_summary}) for clarity. The RL strategy is the core contribution and its evaluation is aligned with the others for a coherent narrative.
    \textcolor{blue}{For each scenario and controller configuration, the simulation was repeated over 10 independent stochastic runs with different random realisations of EV arrivals and system conditions. All reported metrics are presented as the mean $\pm$ standard deviation across these 10 runs.}

        \begin{figure}
        \centering
        \includegraphics[width=\linewidth]
        {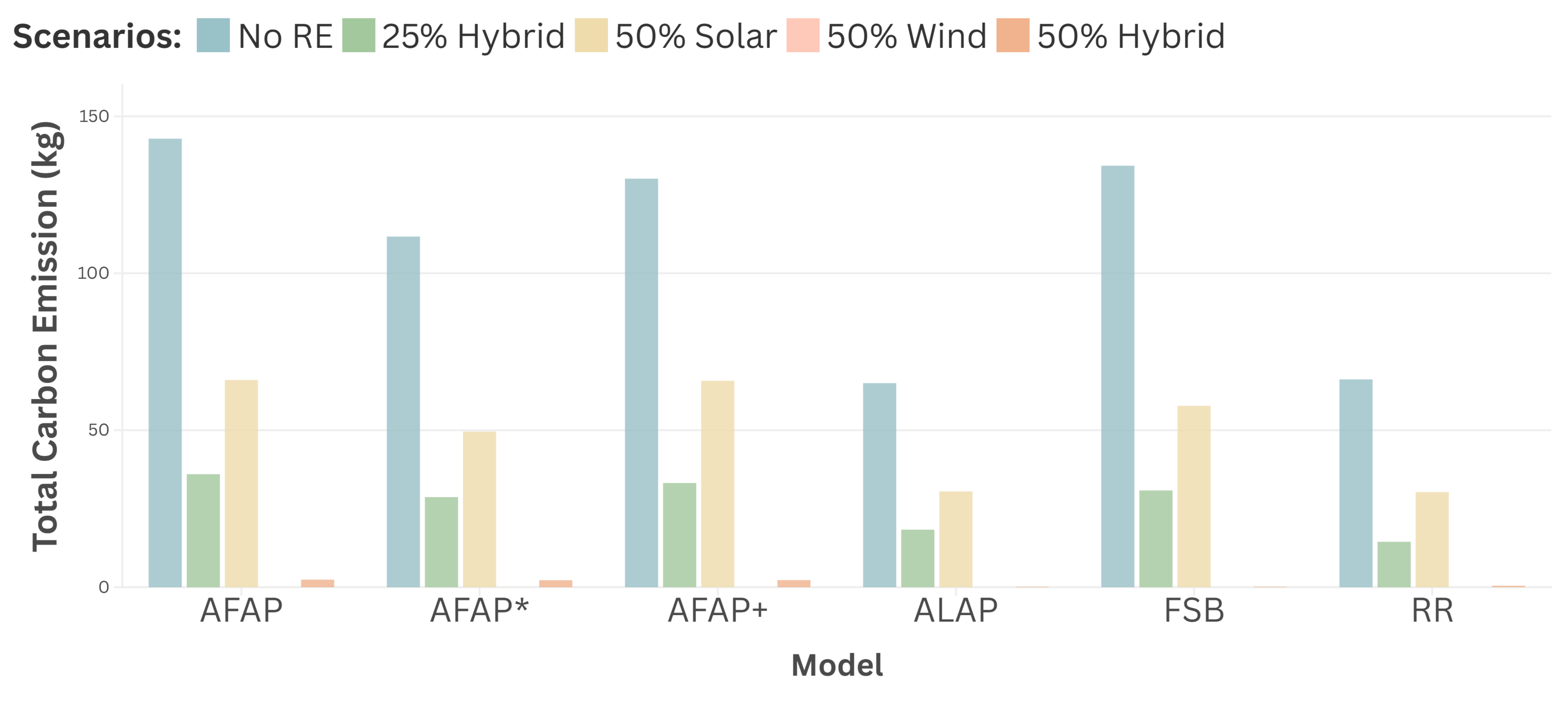}
        \caption{Comparison of baseline strategies. AFAP yields the highest CO$_{2}$
        emissions, whereas ALAP and RR align better with renewable generation and
        achieve lower emissions.}
        \label{fig:co2_emission_baselines}
    \end{figure}

    \subsection{Performance Evaluation of Baseline Charging Strategies}
    The baseline strategies serve as benchmarks, illustrating the extremes of uncoordinated and simplistic coordinated charging approaches. One such baseline is an uncontrolled charging policy, equivalent to the AFAP strategy, in which each EV begins charging at the maximum rate immediately upon plug-in. This greedy approach unsurprisingly achieves near-complete charging for every vehicle by departure, yielding nearly 100\% user satisfaction, but it does so at the cost of severe grid stress.
    \begin{figure*}
        \centering
        \includegraphics[width=\textwidth]{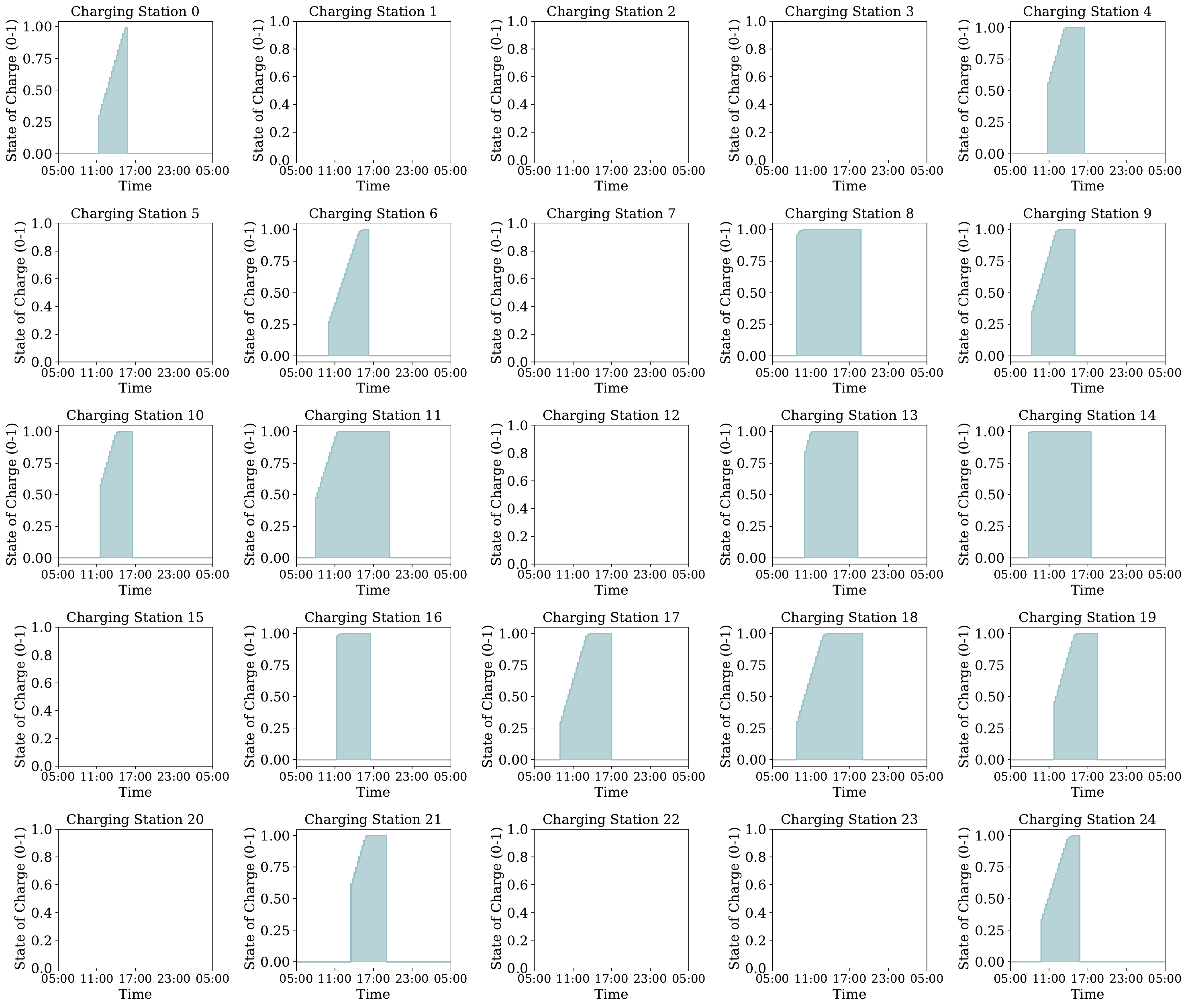}
        \caption{Energy Levels Over Time Across Charging Stations.}
        \label{fig:EV_energy_level}
    \end{figure*}
    
    As shown in Table~\ref{tab:baseline_soc_overload}, the AFAP baseline consistently attains nearly 100\% user satisfaction for departing vehicles, confirming that almost all requested energy is delivered. However, because it disregards network constraints, AFAP frequently causes transformer overloads. In our simulations, overload magnitudes ranged from approximately 442-1328 kWh under high-renewable scenarios to nearly 962 kWh without renewables, far exceeding safe transformer capacity. Fig.~\ref{fig:EV_energy_level} illustrates how AFAP concentrates demand during peak periods, especially around midday when EV presence and solar output both peak, leading to transformer violations. These findings underscore that while AFAP maximises individual charging fulfilment, it is infeasible for real-world deployment due to its severe grid impact.
    \begin{figure}
        \centering
        \includegraphics[width=0.8\linewidth]{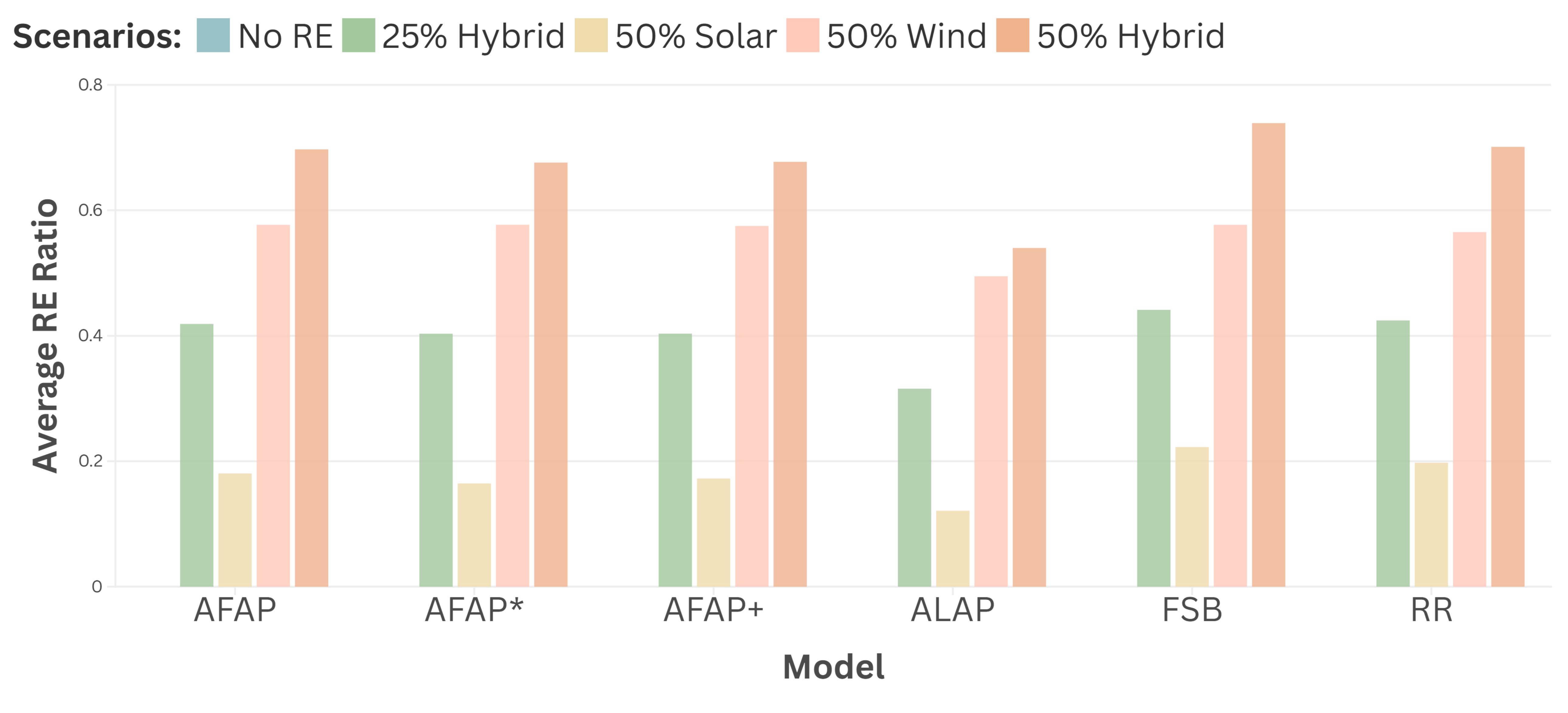}
        \caption{Flexibility of Baseline Charging Strategies Across Scenarios.}
        \label{fig:baseline_re_ratio}
    \end{figure}

    We also evaluated rule-based baselines designed to mitigate these effects. The ALAP heuristic, which postpones charging until just before each EV's departure, reduced transformer overloads to negligible levels across most scenarios. In fact, overloads were effectively 0 kWh under 50\% Solar and 50\% Wind penetration levels, compared to hundreds of kWh with AFAP. However, this improvement comes at the expense of user satisfaction: ALAP achieved only 84-89\% satisfaction, \textcolor{blue}{reflecting its sensitivity to accurately known departure times.}

    Another rule-based baseline, RR, distributes available power evenly across connected EVs. This approach improves fairness and avoids extreme overloads, but user satisfaction dropped further, averaging around 79-84\%. Moreover, RR could not prevent violations when total EV demand exceeded transformer capacity, meaning overloads still occurred under heavy-load conditions.

    Overall, these baselines define the limits of naive and rule-based approaches. AFAP family baselines provide an upper bound for user satisfaction (nearly 100\%) \textcolor{blue}{but represent the worst-case scenario for grid compliance, producing the highest transformer overloads.} ALAP and RR reduce peak demand stress but fail to balance user requirements effectively.

    In terms of carbon emissions and renewable utilisation, baseline strategies also fall short of optimality, as shown in Fig.~\ref{fig:co2_emission_baselines}. AFAP, which charges immediately upon arrival, often coincides with periods of higher grid carbon intensity. Across the no-RE, 25\% Hybrid, and 50\% Solar scenarios, AFAP produces the highest emissions—up to around 143 kg CO$_{2}$ per day. ALAP reduces emissions to about 18–65 kg, while RR typically ranges from 14–66 kg (e.g., 66 kg in no-RE, 14 kg in 25\% Hybrid, 30 kg in 50\% Solar). For example, in the 50\% Solar scenario, AFAP emits 66 kg CO$_{2}$ versus 30 kg for ALAP. In the 50\% Hybrid and 50\% wind scenarios, emissions drop sharply across all strategies; under 50\% wind they are effectively zero (0–0.03 kg). Among the AFAP family of strategies, the FSB consistently produced the highest emissions, as its fixed 10:00 start aligned with periods of rising grid carbon intensity.

    Fig.~\ref{fig:baseline_re_ratio} further illustrates the RE share achieved by each strategy, defined as the proportion of charging demand met by renewable sources. ALAP consistently recorded the lowest RE levels, often falling below 0.2 in the 50\% Solar case, as its delayed-charging behaviour frequently misaligned with periods of high renewable availability. AFAP, AFAP*, and AFAP+ also showed relatively low RE utilisation due to their aggressive early-charging behaviour. In contrast, RR and FSB achieved higher renewable utilisation, with RE shares reaching up to 0.7 under Hybrid conditions. These findings highlight that baseline strategies are unable to simultaneously minimise emissions, exploit renewable availability, and ensure grid safety. Once realistic network constraints are enforced, their limitations become clear: AFAP maximises user satisfaction but produces excessive overloads and emissions, ALAP achieves lower emissions but misses renewable opportunities, and RR and FSB improve renewable utilisation but cannot fully guarantee grid safety. \textcolor{blue}{These observations are consistent with the broader literature: Gan\-esh et al.~\cite{GANESH2022111833} similarly found that fixed-schedule and delayed-charging heuristics reduce peak demand relative to uncontrolled charging but leave emissions performance highly dependent on the coincidence of charging windows with low-carbon generation periods, while C\c{s}eng\"{o}r et al.~\cite{csengor2018optimal} demonstrated that demand-response-aware scheduling substantially reduces peak loads in parking facilities, yet without an explicit emission signal the carbon benefit remains incidental rather than deliberate.}

    \begin{table}
        \setlength{\tabcolsep}{3.5pt}
        \centering
        \caption{Performance of baseline charging strategies across scenarios. Sat
        = user satisfaction; Tr.OL = transformer overload; Hybrid = Wind + Solar.}
        \label{tab:baseline_soc_overload}
        \begin{tabular}{l l c c c}
            \toprule \textbf{Scenario}           & \textbf{Strategy} & \textbf{Sat (\%)} & \textbf{Tr. OL (kWh)} & \textbf{CO$_{2}$ (kg)} \\
            \midrule \multirow{6}{*}{No RE}      & AFAP              & 99.50             & 961.38                & 142.88                 \\
                                                 & ALAP              & 83.69             & 125.81                & 65.00                  \\
                                                 & AFAP*             & 97.89             & 659.80                & 130.13                 \\
                                                 & AFAP+             & 99.49             & 625.06                & 111.69                 \\
                                                 & FSB               & 98.57             & 1069.06               & 134.25                 \\
                                                 & RR                & 79.29             & 132.04                & 66.19                  \\
            \midrule \multirow{6}{*}{25\% Hybrid}   & AFAP              & 100.00            & 1092.51               & 36.00                  \\
                                                 & ALAP              & 84.58             & 1.91                  & 18.33                  \\
                                                 & AFAP*             & 99.30             & 552.89                & 33.19                  \\
                                                 & AFAP+             & 100.00            & 681.01                & 28.70                  \\
                                                 & FSB               & 100.00            & 1328.39               & 30.83                  \\
                                                 & RR                & 78.78             & 157.07                & 14.46                  \\
            \midrule \multirow{6}{*}{50\% Hybrid}   & AFAP              & 100.00            & 1078.71               & 2.41                   \\
                                                 & ALAP              & 84.58             & 1.30                  & 0.17                   \\
                                                 & AFAP*             & 99.30             & 539.93                & 2.26                   \\
                                                 & AFAP+             & 100.00            & 668.09                & 2.22                   \\
                                                 & FSB               & 100.00            & 1313.92               & 0.18                   \\
                                                 & RR                & 78.78             & 150.65                & 0.46                   \\
            \midrule \multirow{6}{*}{50\% Solar} & AFAP              & 100.00            & 441.83                & 66.00                  \\
                                                 & ALAP              & 89.28             & 0.00                  & 30.52                  \\
                                                 & AFAP*             & 100.00            & 267.07                & 65.75                  \\
                                                 & AFAP+             & 100.00            & 248.30                & 49.56                  \\
                                                 & FSB               & 100.00            & 634.22                & 57.78                  \\
                                                 & RR                & 84.27             & 125.19                & 30.30                  \\
            \midrule \multirow{6}{*}{50\% Wind}  & AFAP              & 100.00            & 426.48                & 0.03                   \\
                                                 & ALAP              & 89.28             & 0.00                  & 0.00                   \\
                                                 & AFAP*             & 100.00            & 256.43                & 0.03                   \\
                                                 & AFAP+             & 100.00            & 237.36                & 0.03                   \\
                                                 & FSB               & 100.00            & 617.34                & 0.00                   \\
                                                 & RR                & 84.27             & 121.04                & 0.00                   \\
            \bottomrule
        \end{tabular}
    \end{table}

    \subsection{Model Predictive Control Strategy Evaluation}

    \begin{figure}
        \centering
        \includegraphics[width=0.8\linewidth]{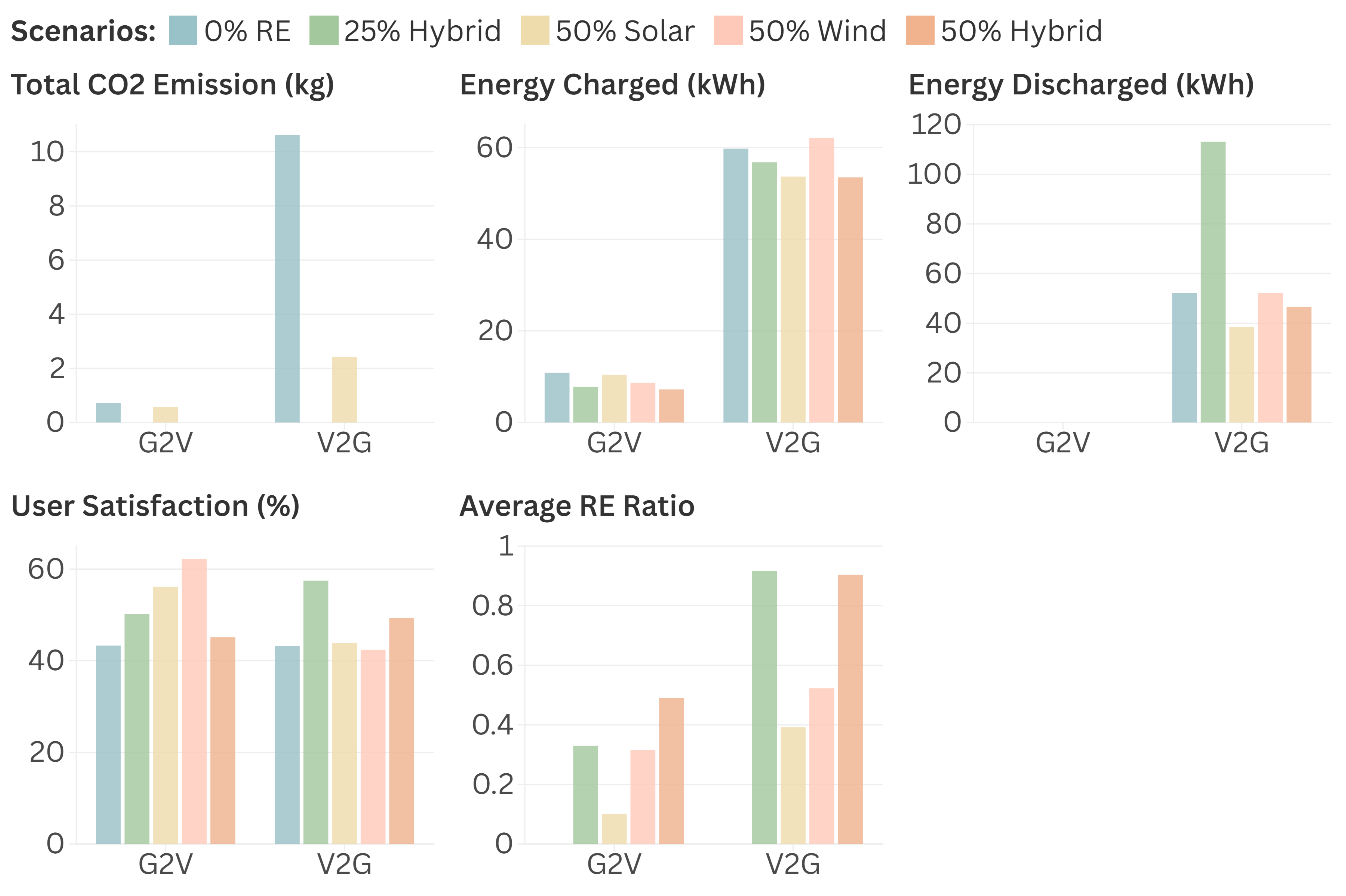}
        \caption{MPC Strategy Performance Across Scenarios.}
        \label{fig:mpc_radar_plots}
    \end{figure}

    \textcolor{blue}{Fig.~\ref{fig:mpc_radar_plots} compares the two MPC strategies, G2V and V2G, across five renewable scenarios using grouped bar charts for total CO$_2$ emissions, energy charged, energy discharged, user satisfaction, and average RE ratio. The figure shows that the main difference between the two controllers is not residual emissions under renewable-rich conditions, but rather renewable utilisation and bidirectional energy cycling. Under 25\% Hybrid, 50\% Wind, and 50\% Hybrid, both controllers reduce total emissions to nearly zero. By contrast, under No RE and 50\% Solar, G2V records lower emissions than V2G, at 0.72~kg versus 10.62~kg and 0.57~kg versus 2.42~kg, respectively. This suggests that the emission benefit of bidirectional operation is scenario-dependent rather than universal, since favourable renewable conditions already allow both MPC variants to operate in a near-zero-emission regime.}
    
        \textcolor{blue}{A clearer separation appears in renewable utilisation. Across all renewable scenarios, V2G achieves a substantially higher average RE ratio than G2V, increasing from 33.0\% to 91.6\% at 25\% Hybrid, from 10.1\% to 39.2\% at 50\% Solar, from 31.5\% to 52.3\% at 50\% Wind, and from 48.9\% to 90.4\% at 50\% Hybrid. This advantage is accompanied by greater battery cycling. G2V performs no discharge in any scenario, whereas V2G discharges between 38.53 and 113.11 kWh across the renewable cases while maintaining charged energy in the range of 53.49--62.14 kWh. These results show that V2G is much more effective at absorbing, shifting, and reusing locally available renewable energy, but it achieves this through substantially more intensive bidirectional operation.}
    
        \textcolor{blue}{User satisfaction shows a more nuanced pattern. In the No RE case, the two controllers are effectively identical, both delivering about 43\% satisfaction. Under single-source renewable supply, G2V remains more user-oriented, reaching 56.13\% versus 43.86\% in the 50\% Solar case and 62.16\% versus 42.38\% in the 50\% Wind case. Under hybrid supply, however, V2G outperforms G2V, increasing satisfaction from 50.21\% to 57.46\% at 25\% Hybrid and from 45.13\% to 49.30\% at 50\% Hybrid. This suggests that the effect of bidirectional flexibility on charging fulfilment depends strongly on the renewable mix. When only one renewable source is available, additional cycling can reduce charging completion. When wind and solar are combined, V2G can use that flexibility more effectively without the same penalty to user satisfaction.}

    \textcolor{blue}{These patterns are broadly consistent with findings reported in the literature. Koufakis et al.~\cite{8710609} similarly showed that bidirectional coordination can improve renewable self-consumption relative to unidirectional charging, which is strongly reflected here in the substantially higher RE ratios achieved by V2G. Meenakumar et al.~\cite{9242538} also found that the value of V2G depends on balancing the operational benefits of discharging against battery wear, a trade-off that is evident in the much larger charged and discharged energy volumes observed for V2G in Fig.~\ref{fig:mpc_radar_plots}. The near-identical satisfaction of the two modes in the No RE case further suggests that, when renewable availability is absent, the additional flexibility of bidirectional control offers only limited user-level benefit, which is consistent with the general multi-objective charging literature~\cite{qureshi2024multiobjective}.}

    \subsection {\textcolor{blue}{Reinforcement Learning} Strategy Evaluation}
    To evaluate the effectiveness of our emission-aware RL agent, we conducted experiments under five distinct scenarios: No RE, 50\% Wind, 50\% Solar, 25\% Hybrid, and 50\% Hybrid. Each scenario was simulated for 10 independent runs, and the results reported below are averages across these runs to ensure robustness.

    Performance was assessed using key metrics aligned with the design objectives: CO$_2$ emissions, renewable utilisation (RE ratio), user satisfaction (Sat), transformer overload, and energy throughput (charging and discharging). The reward function applied during training reflected these priorities with the following weighting:

    \begin{itemize}
        \item Discharging penalty: $W_{D}=50$
        \item CO$_2$ emissions penalty: $W_{\text{CO}2}=5$
        \item User satisfaction penalty (triggered if SoC $<0.8$): $W_{\text{sat}}=50$
    \end{itemize}

    These values, illustrated in Fig.~\ref{fig:flow_chart}, \textcolor{blue! }{were determined through preliminary experimental tuning and are further justified by the weight sensitivity analysis in Section~\ref{sec:ablation}. }They strongly discourage unnecessary V2G actions, prioritise low-carbon charging windows, and preserve service quality for EV drivers.

    \begin{figure}
        \centering
       \includegraphics[width=0.7\linewidth]{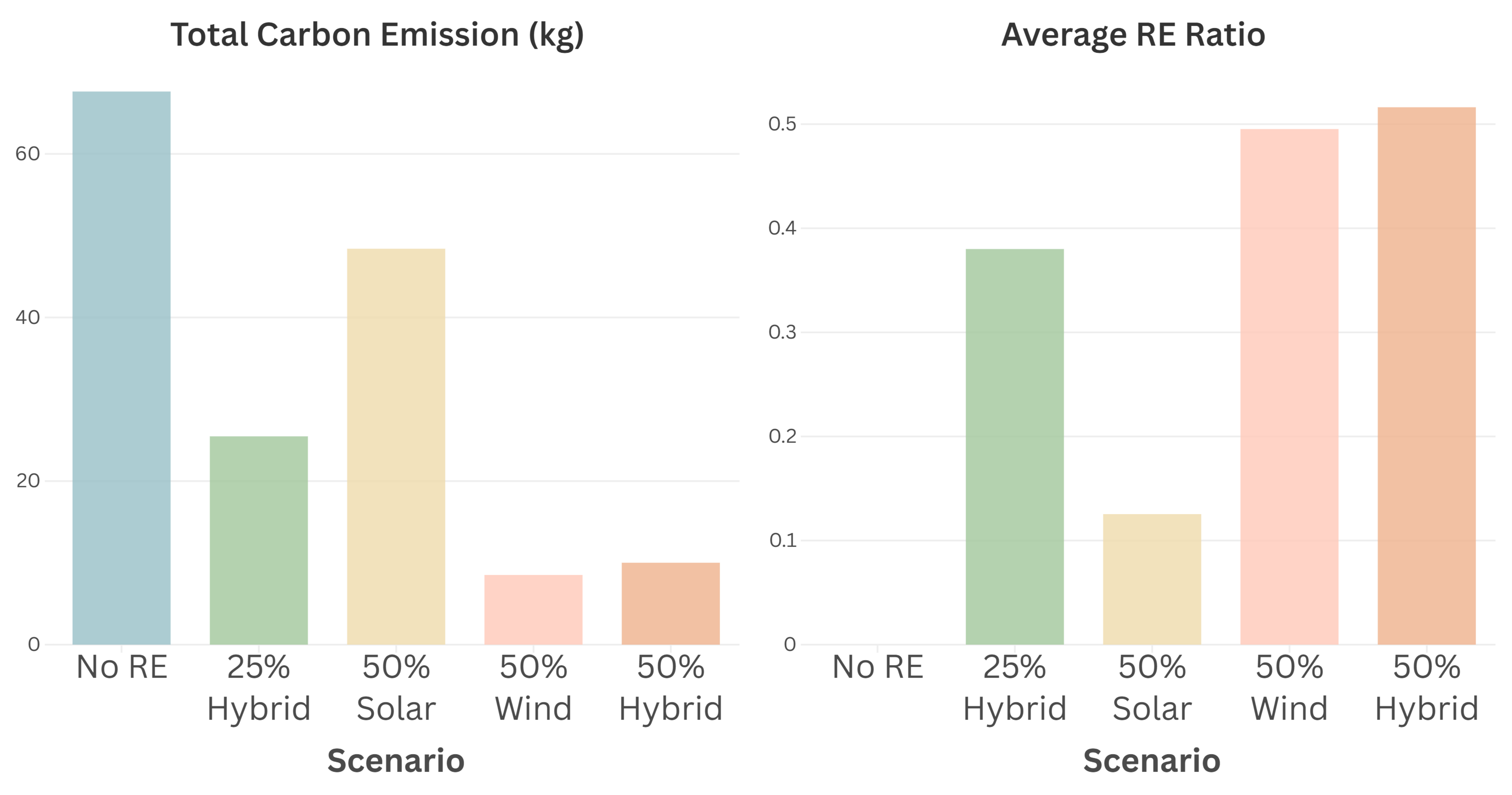}
        \caption{CO$_{2}$ Emissions of RL Strategies Across Scenarios.}
        \label{fig:rl_co2_re}
    \end{figure}

    \textcolor{blue! }{As shown in Fig.~\ref{fig:rl_co2_re}, the integrated results highlight a clear relationship between carbon reduction and renewable utilisation across scenarios. In the left panel, the RL agent achieves the lowest total CO$_2$ emissions under 50\% Wind (8.52 kg) and 50\% Hybrid (10.02 kg), while emissions rise to 25.47 kg under 25\% Hybrid and 48.41 kg under 50\% Solar, reaching the highest level of 67.62 kg in the No RE case. This pattern shows that the agent benefits most from scenarios with stronger or more temporally distributed renewable supply, particularly wind, which allows charging to be shifted towards lower-carbon periods.}

    \textcolor{blue! }{The right panel shows the corresponding average renewable energy (RE) ratio, defined as the proportion of charging energy supplied by renewables. The RL agent attains the highest RE ratios under 50\% Hybrid (0.52) and 50\% Wind (0.50), followed by 25\% Hybrid (0.38), while utilisation remains much lower under 50\% Solar (0.13) and is zero in the No RE case. Taken together, these results show that the proposed state design allows the agent to identify and exploit renewable availability when present, and that higher renewable utilisation is closely associated with lower charging-related emissions.}

    Table~\ref{tab:rl_summary} presents a consolidated view of the RL agent's performance across all scenarios. Despite prioritising emissions and grid compliance, the model maintains moderate user satisfaction, with user satisfaction ranging from 49.28\% to 55.60\%. While lower than the near-100\% scores of greedy baselines, this performance is comparable to discharging-based V2G models, which similarly trade off individual user satisfaction for broader system benefits like peak shaving or emission reduction.

\begin{figure}
        \centering
        \includegraphics[width=0.7\linewidth]{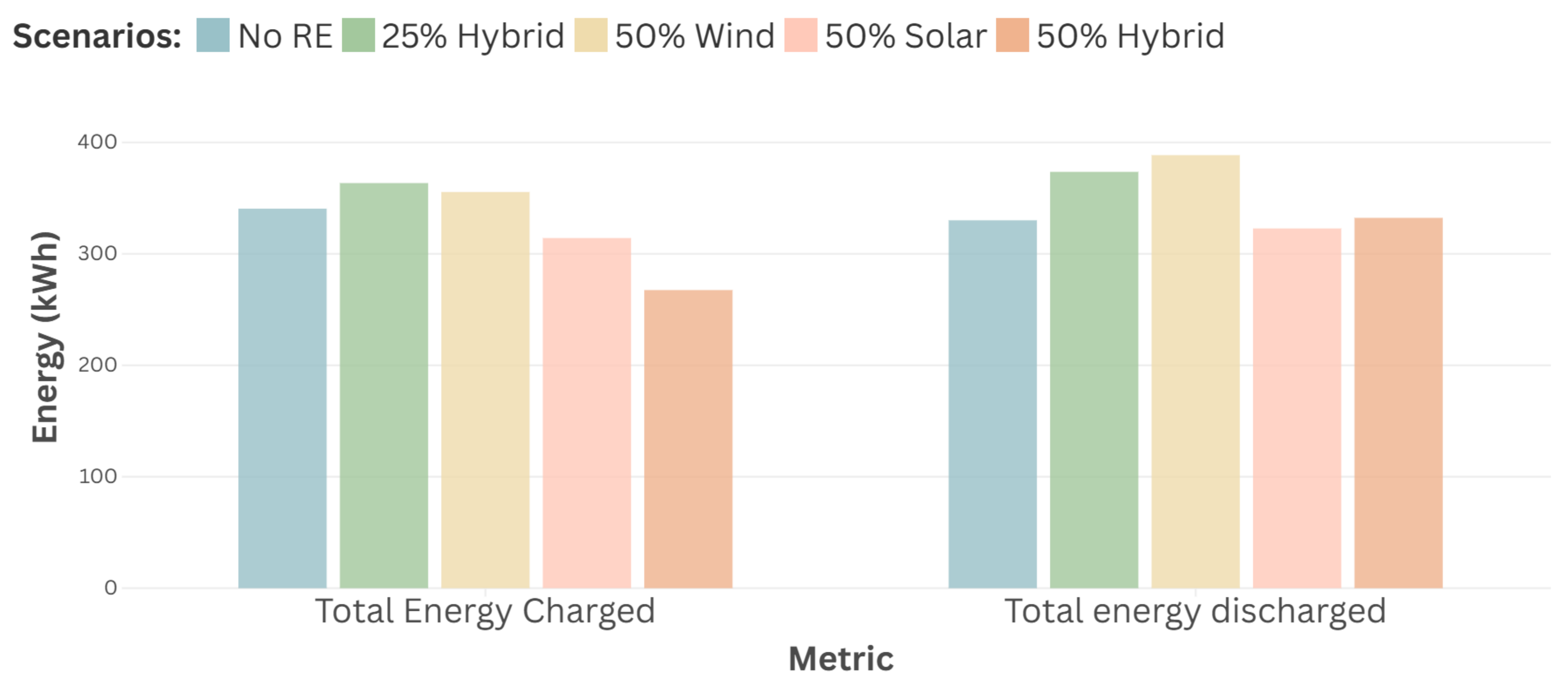}
        \caption{Energy Charged and Discharged by RL Strategies Across Scenarios.}
        \label{fig:rl_charge_discharge}
    \end{figure}

    Additionally, the RL agent achieves strong energy throughput, charging and discharging over 300 kWh in all settings (Fig.~\ref{fig:rl_charge_discharge}). This confirms its dual functionality: fulfilling EV demand and supporting system-level flexibility via V2G, guided by our reward design.


    Crucially, the RL agent strictly adheres to grid constraints. Transformer overload remains below 7 kWh in all scenarios, and is as low as 0.52 kWh in the No RE case. This contrasts sharply with heuristic baselines, which often report lower emissions or higher satisfaction at the cost of infeasible overloads. Here, compliance is baked into the policy's behaviour, not enforced post hoc.

    \begin{table*}
        [t]
        \centering
        \caption{RL Strategy Metrics Averaged Across 10 Episodes per Scenario}
        \label{tab:rl_summary}
        \begin{tabular}{lcccccc}
            \toprule \textbf{Scenario} & \textbf{CO$_{2}$ (kg)} & \textbf{RE Ratio} & \textbf{Sat (\%)} & \textbf{Transformer Overload (kWh)} & \textbf{Charged (kWh)} & \textbf{Discharged (kWh)} \\
            \midrule No RE             & 67.62                 & 0.00              & 55.60             & 0.52                                & 340.58                 & 330.13                    \\
            50\% Wind                  & 8.52                  & 0.50              & 52.00             & 6.83                                & 355.53                 & 388.61                    \\
            50\% Solar                 & 48.41                 & 0.13              & 51.29             & 5.94                                & 314.21                 & 322.83                    \\
            25\% Hybrid            & 25.47                 & 0.38              & 52.69             & 1.62                                & 363.54                 & 373.60                    \\
            50\% Hybrid            & 10.02                 & 0.52              & 49.28             & 3.77                                & 267.53                 & 332.33                    \\
            \bottomrule
        \end{tabular}
    \end{table*}

    \begin{table}
        \centering
        \caption{Carbon Intensity (CI) Comparison between Emission-Aware SAC and
        PDN Models}
        \label{tab:ci_comparison}
        \begin{tabular}{lcc}
            \toprule \textbf{Model}                           & \textbf{Scenario} & \textbf{CI (gCO$_{2}$/kWh)} \\
            \midrule \multirow{5}{*}{Emission-Aware SAC}      & No RE             & 198.54                      \\
                                                              & 50\% Wind         & 23.96                       \\
                                                              & 50\% Solar        & 154.07                      \\
                                                              & 25\% Hybrid   & 70.06                       \\
                                                              & 50\% Hybrid   & 37.45                       \\
            \midrule \multirow{3}{*}{PDN~\cite{qiu2024graph}} & Case 1            & 137.32                      \\
                                                              & Case 2            & 100.11                      \\
                                                              & Case 3            & 95.75                       \\
            \bottomrule
        \end{tabular}
    \end{table}

    \textcolor{blue}{To allow fair comparison with external work, we also report the carbon intensity (CI)} \textcolor{blue!60! black}{that is defined and formulated in EirGrid~\cite{smartgrid-co2}}, \textcolor{blue}{calculated as:}
    
\begin{equation}
    CI = \dfrac{\text{Total CO}_{2}\text{ Emission} (kg)}{\text{Total Energy Charged (kWh)}} \quad [\text{kgCO}_{2}\text{/kWh}]
    \label{eq:CI_calculation}
\end{equation}

    This normalisation enables direct benchmarking against models that operate under different scales or network conditions.

    Table~\ref{tab:ci_comparison} compares our CI values with those reported for the PDN model~\cite{qiu2024graph}. \textcolor{blue! }{For ease of comparison and to preserve numerical precision in presentation, our CI values are reported here in $\text{gCO}_{2}\text{/kWh}$ rather than $\text{kgCO}_{2}\text{/kWh}$.} The Emission-Aware SAC achieves substantially lower CI in renewable-integrated settings, such as 23.96~gCO$_{2}$/kWh in the 50\% Wind case and 37.45~gCO$_{2}$/kWh in the 50\% Hybrid case, compared with 95–137~gCO$_{2}$/kWh reported for PDN. These results highlight the strong advantage of our approach when high RE penetration is available, demonstrating that the RL policy can align charging with clean energy supply far more effectively than the external baseline. Although CI remains higher in No-RE and Solar settings, these scenarios are inherently constrained by limited renewable availability, reinforcing the importance of carbon-aware scheduling under renewable integration.

    Compared with heuristic models and the V2G baseline, the RL policy offers a more holistic and deployable solution for real-time smart charging. Relative to both internal references and the external PDN model, it achieves substantially lower carbon intensity while maintaining system feasibility.

    \textcolor{blue}{These results contextualise the contribution within the broader RL literature. Rossi et al.~\cite{rossi2025smart} demonstrated that model-free RL agents can learn effective charging policies without explicit forecasts, but their formulation did not incorporate a carbon intensity signal; the gap in emission performance between that work and the present study illustrates the direct benefit of embedding a time-varying CO$_2$ penalty in the reward. The review by Qiu et al.~\cite{QIU2023113052} notes that most existing RL charging agents optimise cost or load rather than emissions, and that few simulators provide the transformer-level and V2G constraints needed for realistic evaluation -- both limitations addressed by the EV2Gym-based framework used here. Furthermore, the CI values achieved by the emission-aware SAC (as low as 23.96~gCO$_2$/kWh under 50\% wind) compare favourably against the graph-based PDN model of Qiu et al.~\cite{qiu2024graph}, which reported CI values of 95--137~gCO$_2$/kWh despite jointly optimising routing and charging. The margin is attributable to the direct inclusion of a dynamic carbon intensity forecast in the state representation, enabling the SAC policy to time-shift charging to coincide with low-emission grid periods -- a capability not present in the PDN model's objective formulation.}
    
    \subsection{\textcolor{blue! }{Training Behaviour Analysis}}

        \begin{figure}
         \centering
        \includegraphics[width=0.7\linewidth]{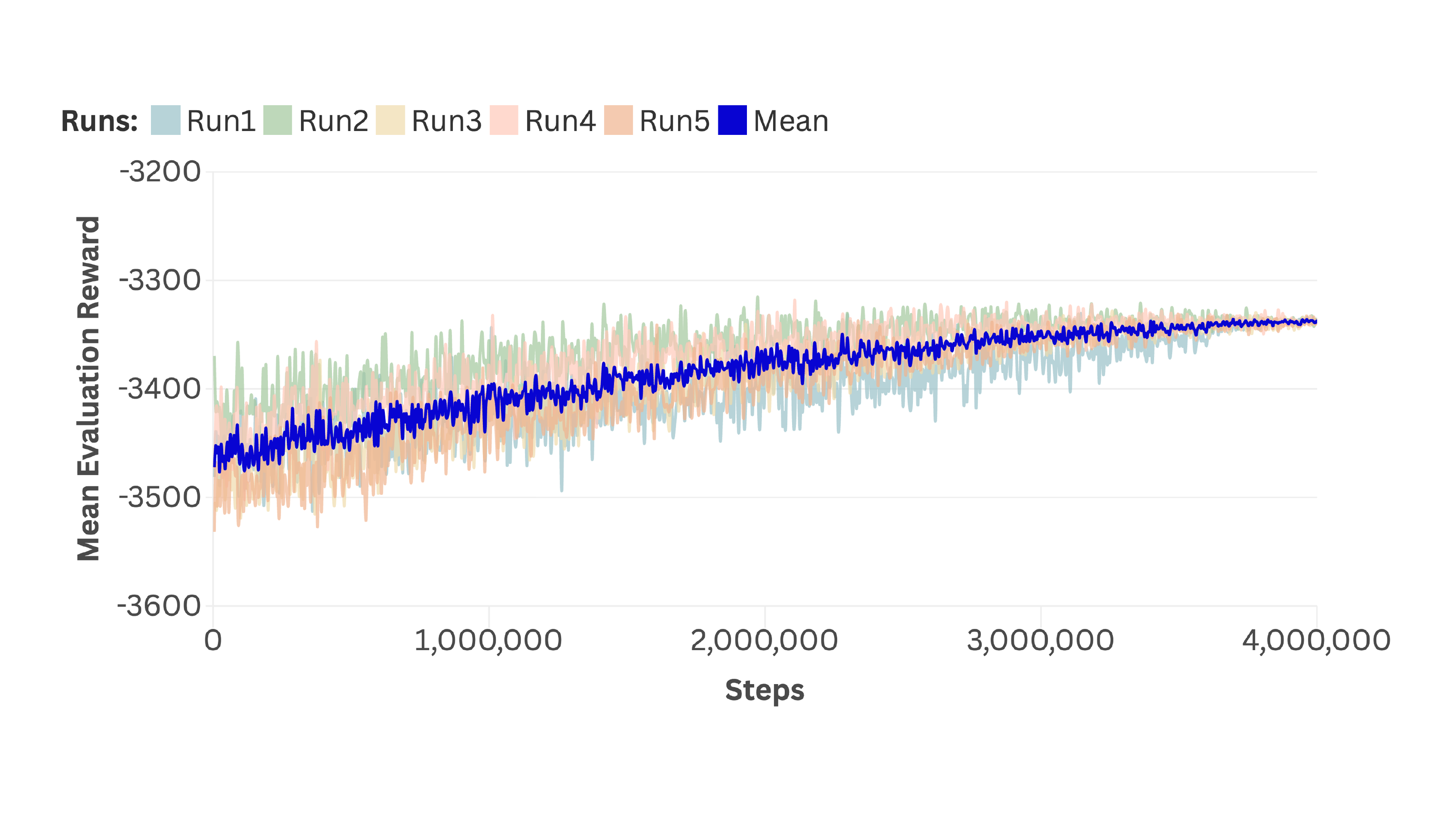}
        \caption{\textcolor{blue! }{Evaluation reward trajectories of the SAC agent over 4 million training steps across five independent runs.}}
        \label{fig:run_mean_reward}
    \end{figure}

    \textcolor{blue}{Fig.~\ref{fig:run_mean_reward} shows the evaluation reward trajectories for the five runs. Although the curves remain noisy, which is expected in a stochastic charging environment, they follow a consistent upward trend throughout training. The mean evaluation reward improves substantially during the early and intermediate phases of training, then increases more slowly in the final stage. This suggests practical convergence rather than full saturation: the policy continues to improve, but the marginal gains become smaller toward the end of training. The relatively tight grouping of the five runs in the later stage also indicates that the learning behaviour is reproducible and not driven by a single favourable seed.}
        
    \begin{figure}
         \centering
        \includegraphics[width=0.9\linewidth]{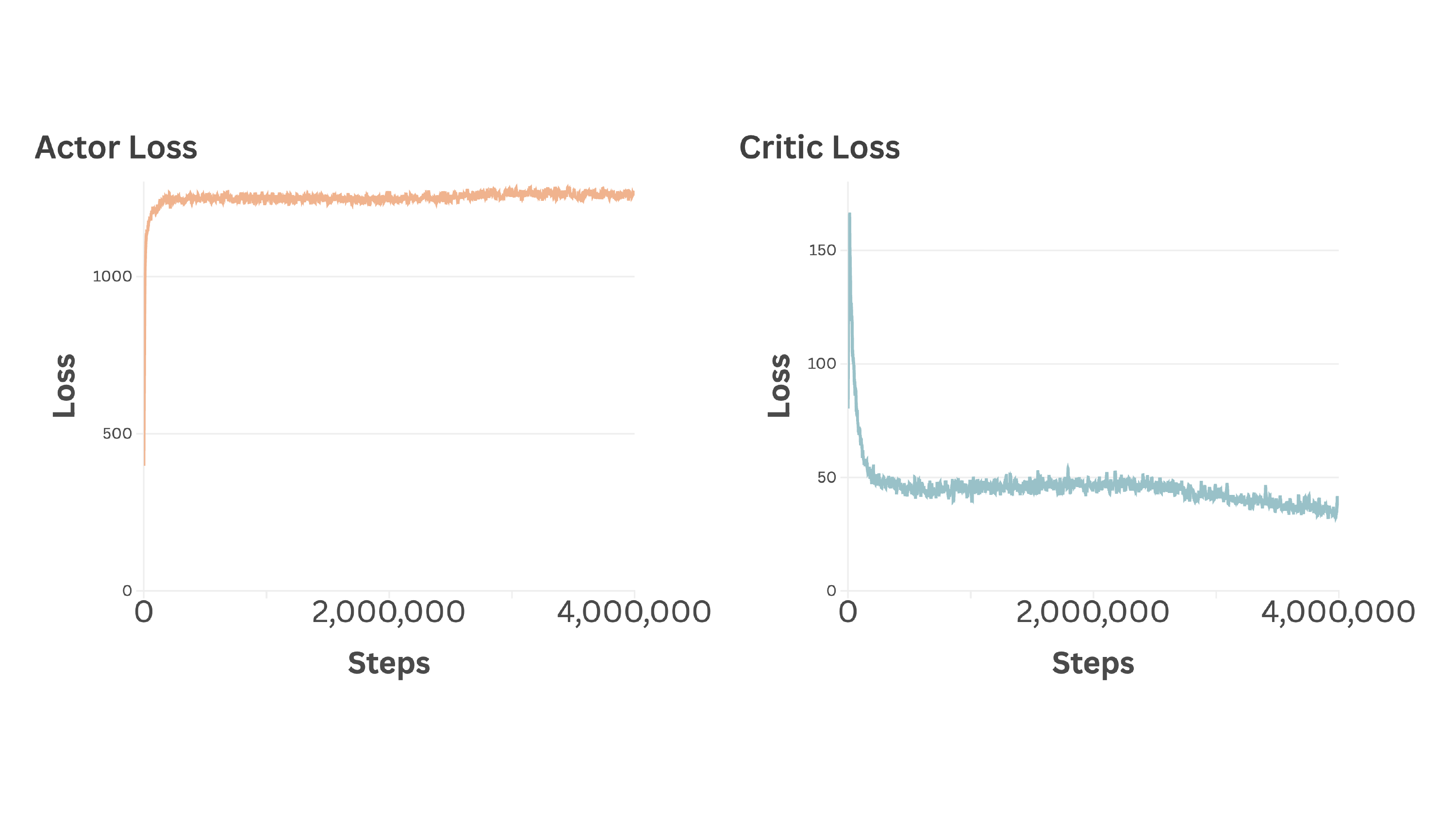}
        \caption{\textcolor{blue}{Actor and critic loss curves from a representative SAC training run over 4 million steps.}}
        \label{fig:actor_critic_loss}
    \end{figure}

    \textcolor{blue! }{Fig.~\ref{fig:actor_critic_loss} provides supporting evidence from the optimisation process. The actor loss rises sharply at the start of training and then remains within a bounded range, with only mild drift over the remaining steps. The critic loss drops quickly from its initial level, stays comparatively low, and decreases slightly again near the end of training. These traces do not by themselves establish convergence, but they are consistent with a stable optimisation process and show no sign of late-stage divergence, collapse, or uncontrolled oscillation. Convergence is therefore assessed primarily from the evaluation reward trends, with the loss curves used as secondary evidence of training stability.}

    \begin{table}[t]
    \centering
    \caption{Computational Information}
    \label{tab:computational_cost}
    \begin{tabular}{p{0.40\linewidth} p{0.52\linewidth}}
    \toprule
    \textbf{Item} & \textbf{Value} \\
    \midrule
    Algorithm & SAC \\
    Training steps & 4,000,000 \\
    Hardware & Apple MacBook Pro (Apple M4 Pro, 14-core CPU, 48 GB memory) \\
    Device & CPU \\
    Training throughput & 103 steps/s \\
    Training time per run & 10 h 45 m 49 s \\
    Inference time per step & 0.116 $\pm$ 0.029 ms \\
    Measured decision steps & 5,740 \\
    \bottomrule
    \end{tabular}
    \end{table}

    \textcolor{blue! }{The computational information of the proposed RL controller is summarised in Table~\ref{tab:computational_cost}. Training the SAC agent for 4 million steps required an average wall-clock time of 10 h 45 m 49 s per run on an Apple MacBook Pro with an Apple M4 Pro chip, a 14-core CPU, and 48 GB of memory, using CPU-only training. This corresponds to an average throughput of approximately 103 steps per second. After training, policy execution was lightweight, with an average inference time of 0.116 $\pm$ 0.029 ms per decision step, measured over 5,740 post-warm-up decision steps. The main computational burden therefore lies in offline training rather than online control. Overall, the reward trajectories, loss curves, and timing results indicate that the proposed SAC controller learns stably, approaches practical convergence within the chosen training horizon, and remains efficient for real-time deployment in the considered EV charging setting.}
    
    \subsection{Ablation and Sensitivity Analysis}
    \label{sec:ablation}
    
    \textcolor{blue! }{To assess the adequacy of the reward design, examine renewable dependence, and evaluate robustness, four supplementary analyses were conducted. First, a reward ablation study was performed by removing one reward component at a time from the full formulation. Second, the sensitivity of the learned policy to the carbon penalty weight $W_{CO_{2}}$ was evaluated relative to the selected baseline $W_{CO_{2}}=5$. Third, the selected policy was tested across a wider renewable penetration sweep under Solar, Wind, and Hybrid renewable settings. Fourth, the policy was evaluated under low, medium, and high charging-demand conditions to examine robustness to demand variability. Together, these experiments clarify how the reward terms shape controller behaviour and how stable the proposed approach remains as system conditions change. In each experiment, each RL configuration was evaluated over 30 independent stochastic episodes per scenario.}
    
    \subsubsection{\textcolor{blue! }{Reward Function Ablation Study}}
    \label{subsec:reward-abl}
    
    \begin{figure}
        \centering
        \includegraphics[width=0.9\linewidth]{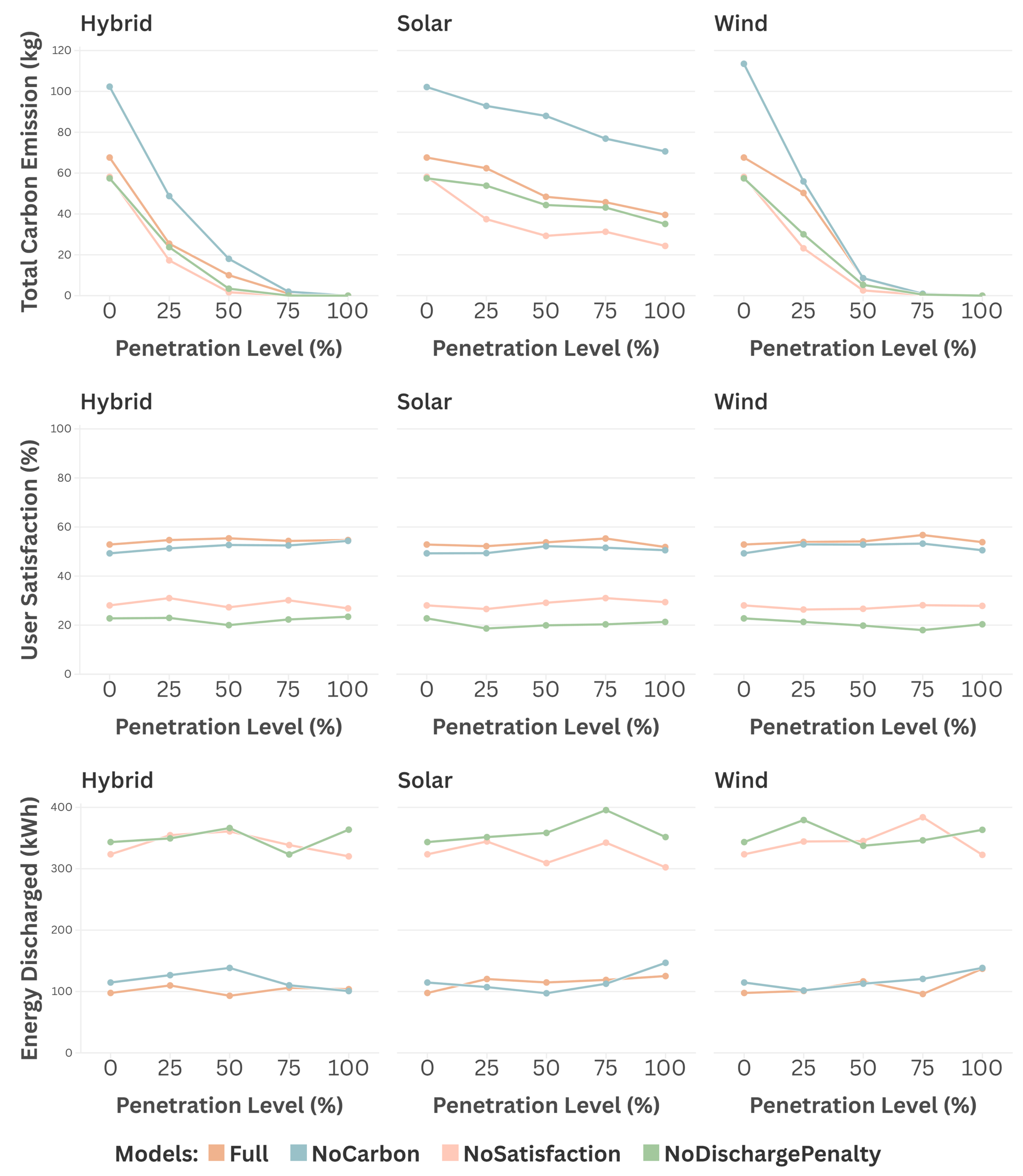}
        \caption{\textcolor{blue! }{Reward component ablation under different renewable settings and penetration levels.}}
        \label{fig:groupA_ablation_reward_function}
    \end{figure}

    \textcolor{blue! }{Fig.\ref{fig:groupA_ablation_reward_function} compares the Full, NoCarbon($W_{CO_{2}}=0$), NoSatisfaction($P_{t}^{Sat}=0$), and NoDischarge($W_{D}=0$), reward variants across Hybrid, Solar, and Wind settings over increasing renewable penetration. The results show that the full reward gives the most balanced behaviour, while removing individual terms pushes the controller toward narrower objectives. In particular, removing the carbon term generally increases total carbon emission, especially at low to medium penetration in the Hybrid and Wind cases, with 48.75~kg versus 25.47~kg at 25\% Hybrid and 55.94~kg versus 50.28~kg at 25\% Wind. This confirms that the explicit carbon term is a real driver of low-emission scheduling rather than a redundant proxy for renewable availability.}

    \textcolor{blue! }{Removing the satisfaction term often reduces emissions further, but only by sacrificing charging completion, with satisfaction falling from 53.75\% to 29.11\% in the 50\% Solar case. Removing the discharge penalty produces a different trade-off: emissions may fall, but discharged energy rises sharply, indicating more aggressive V2G cycling, for example, from 93.07~kWh to 366.32~kWh at 50\% Hybrid. Taken together, the ablation study shows that the carbon term reduces emissions, the satisfaction term prevents underserving EVs, and the discharge penalty keeps V2G behaviour operationally realistic.}

    \subsubsection{\textcolor{blue! }{CO$_{2}$ Weight Sensitivity Analysis}}

    \begin{figure}
         \centering
        \includegraphics[width=0.9\linewidth]{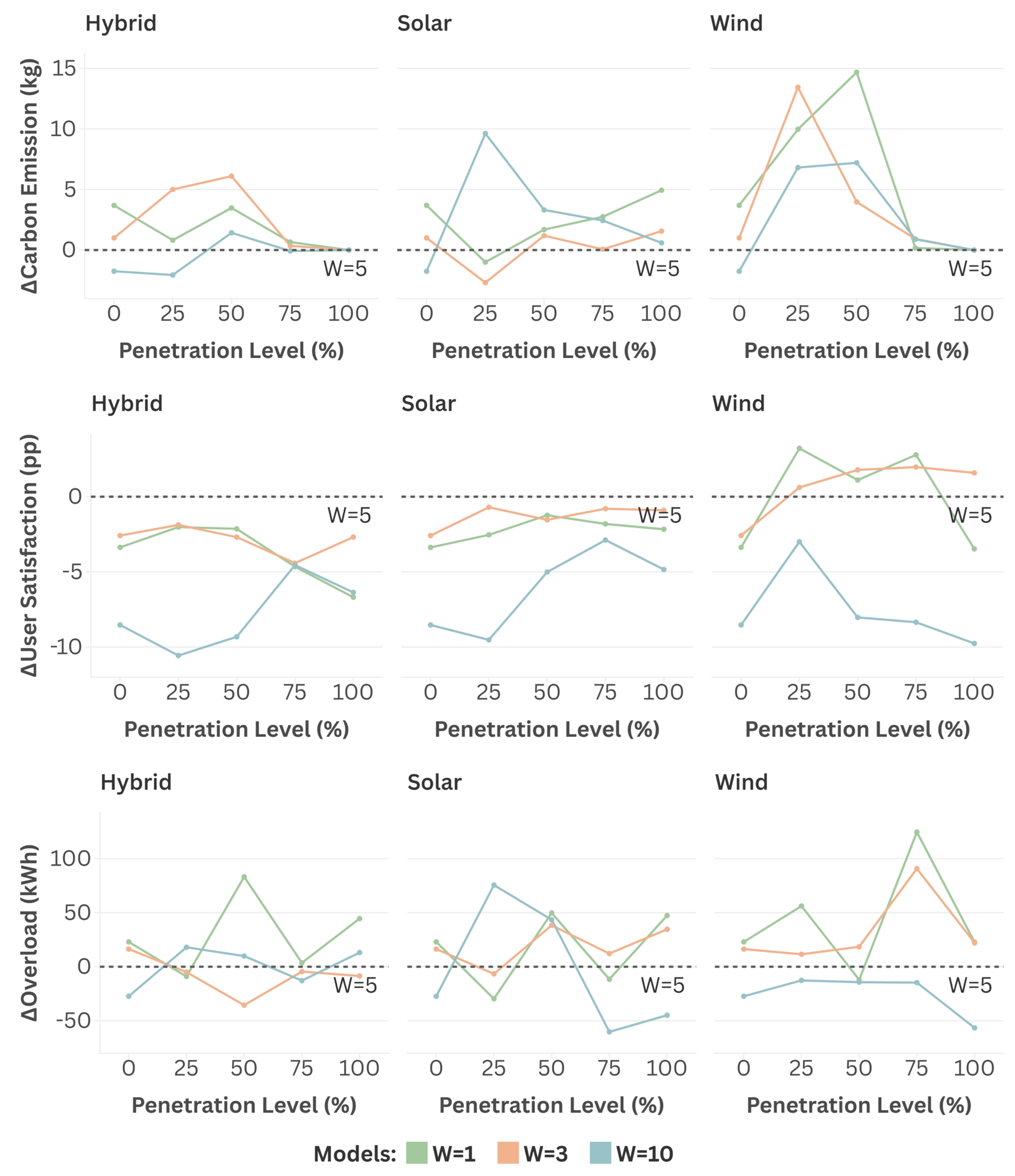}
        \caption{\textcolor{blue! }{Sensitivity of policy performance to the carbon reward weight relative to the $W=5$ baseline.}}
        \label{fig:groupB_ablation_weight}
    \end{figure}

    \textcolor{blue! }{Fig.\ref{fig:groupB_ablation_weight} shows the change in carbon emission, user satisfaction, and transformer overload for $W_{CO_{2}}=1$, $3$, and $10$ relative to the $W_{CO_{2}}=5$ baseline. No alternative weight consistently outperforms the selected baseline. Lower carbon weights, especially $W_{CO_{2}}=1$, generally increase emissions in the Hybrid and Wind settings at low to medium penetration, with 49.38~kgCO$_2$ at 25\% Hybrid and 50.22~kgCO$_2$ at 25\% Wind. This indicates that a weak carbon signal is not sufficient to shift charging reliably away from high-emission periods.
    Increasing the weight to $W_{CO_{2}}=10$ does not yield a uniform gain. In some cases, it lowers emissions, but this is typically accompanied by a larger loss in user satisfaction or less stable overload behaviour, such as 10.5\% at 25\% Hybrid, 9.52\% at 25\% Solar, and 8.34\% at 75\% Wind. Overall, $W_{CO_{2}}=5$ emerges as the best-balanced choice: smaller weights weaken decarbonisation, while larger weights over-prioritise carbon reduction at the expense of other objectives.}

    \subsubsection{\textcolor{blue! }{Robustness Across Renewable Penetration Levels}}

    \begin{figure}
         \centering
        \includegraphics[width=0.7\linewidth]{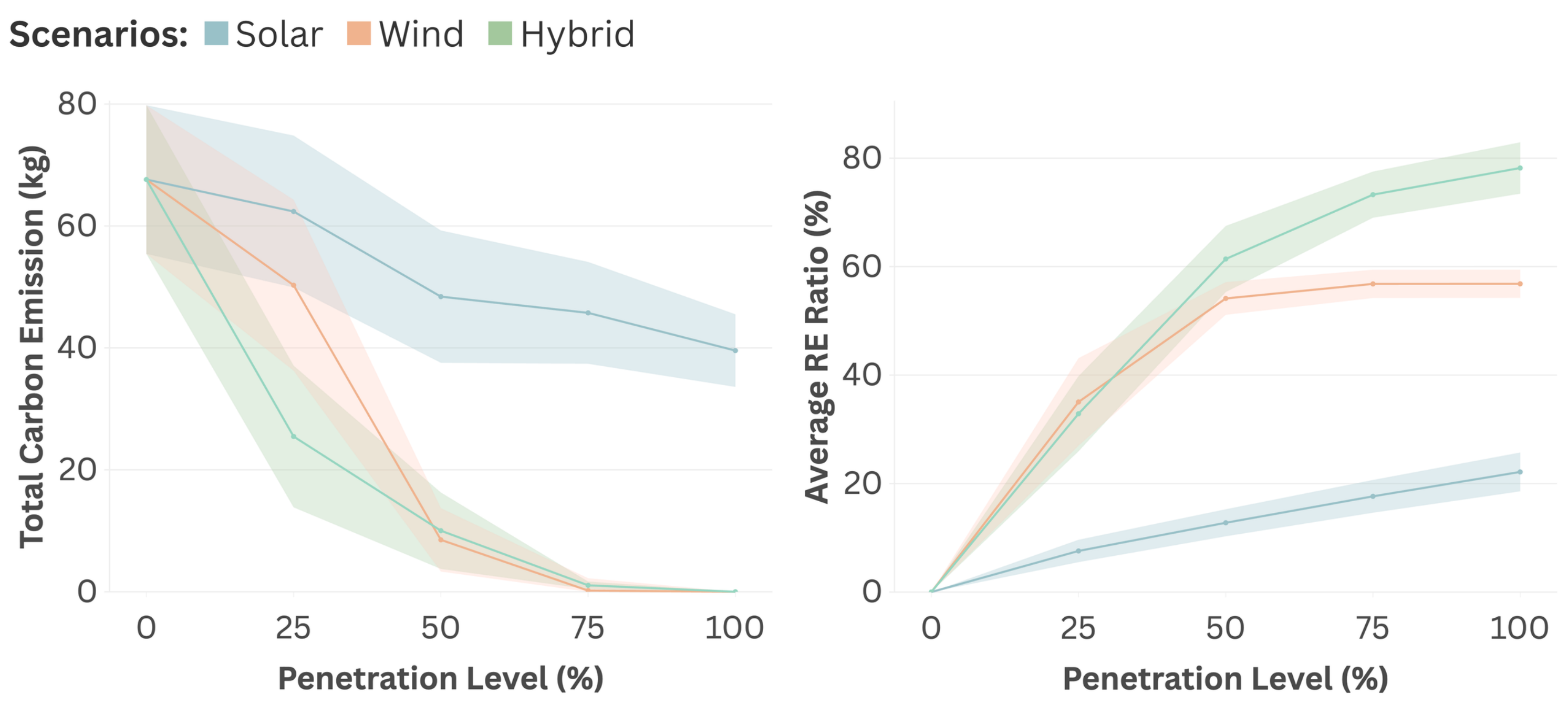}
        \caption{\textcolor{blue! }{Effect of renewable penetration on renewable utilisation and carbon emission for the selected RL policy.}}
        \label{fig:groupC_renewable_penetration}
    \end{figure}

    \textcolor{blue! }{Fig.\ref{fig:groupC_renewable_penetration} evaluates the selected RL policy across a wider renewable-penetration sweep under Solar, Wind, and Hybrid supply. In all three cases, carbon emission decreases as renewable penetration rises, but the rate of improvement depends strongly on the renewable mix. Wind and Hybrid settings show much steeper emission reductions than Solar, with emissions falling from 104.13~kg to 0~kg in the Wind and Hybrid cases, while the Solar case declines more gradually from 104.13~kg to 66.19~kg. This shows that renewable structure matters as much as renewable quantity.}

    \textcolor{blue! }{The renewable-utilisation curves show the same pattern. The average renewable energy ratio rises most strongly in the Wind setting, reaching 56.83\%, followed by the Hybrid case at 78.19\%, while the Solar case remains lower at 22.14\%. The shaded bands also narrow at higher penetration levels, especially under Wind and Hybrid supply, indicating that the learned policy becomes more stable as more clean energy is available.}

    \subsubsection{\textcolor{blue! }{Sensitivity to Charging Demand Variability}}

    \begin{figure}
         \centering
        \includegraphics[width=0.7\linewidth]{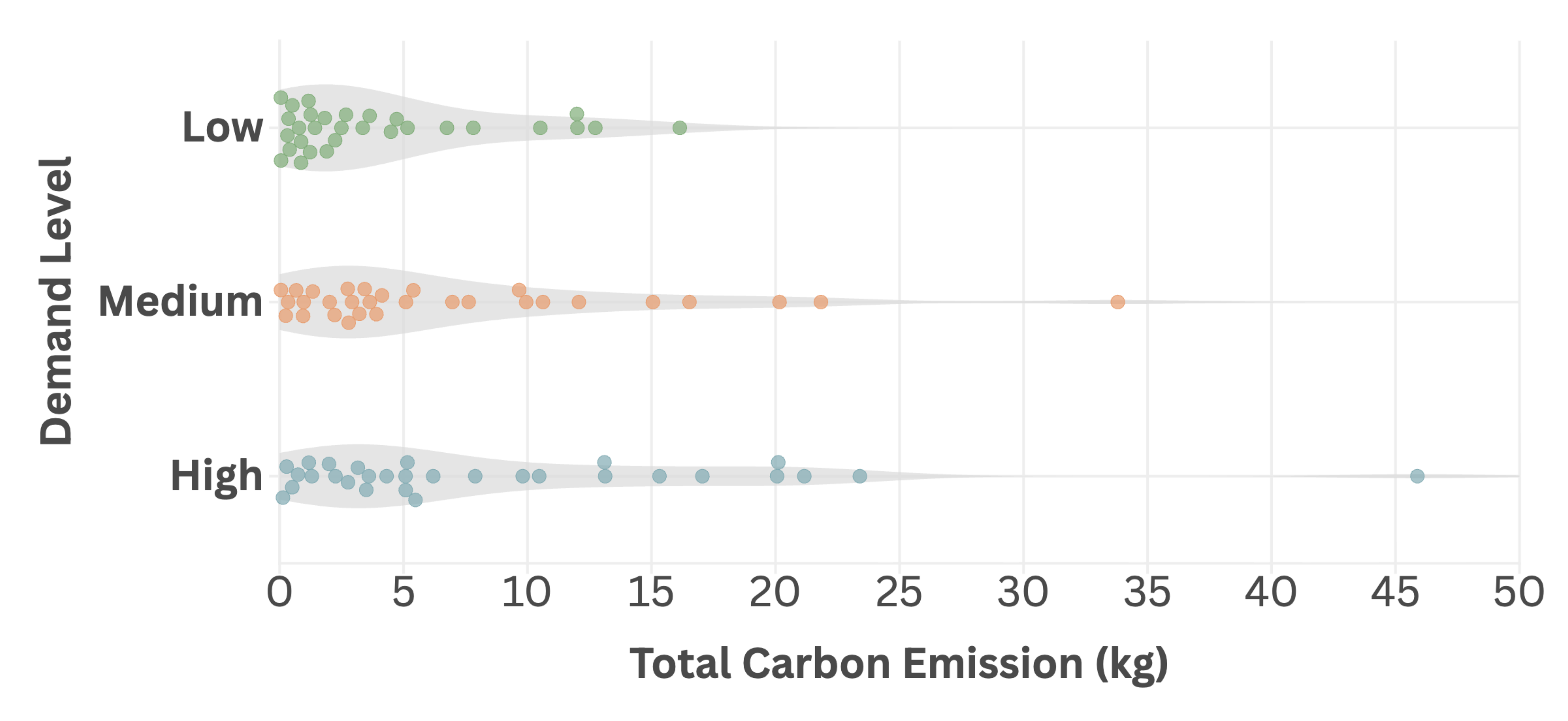}
        \caption{\textcolor{blue! }{Distribution of carbon emission under different charging demand levels.}}
        \label{fig:groupD-carbon}
    \end{figure}

\begin{figure}
         \centering
        \includegraphics[width=0.8\linewidth]{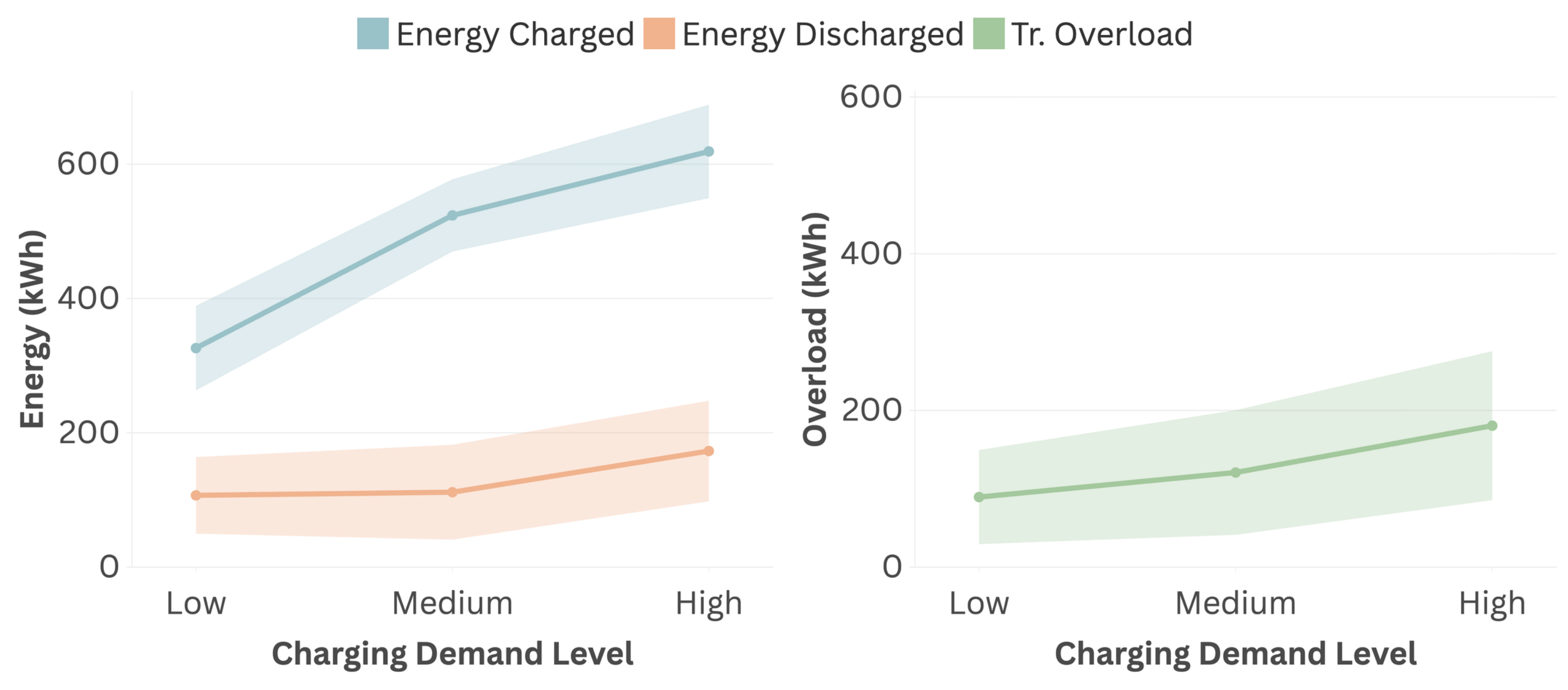}
        \caption{\textcolor{blue! }{Operational response of the selected RL policy under different charging demand levels.}}
        \label{fig:groupD_energy}
    \end{figure}

    \textcolor{blue! }{Figs.~\ref{fig:groupD-carbon} and \ref{fig:groupD_energy} examine the response of the selected RL policy under low-, medium-, and high-demand conditions in the fixed 50\% Hybrid setting. Fig.~\ref{fig:groupD-carbon} shows a clear rightward shift in the distribution of carbon emission as demand increases. The mean rises from 3.99~kgCO$_2$ under low demand to 7.01~kgCO$_2$ under medium demand and 9.0~kgCO$_2$ under high demand, while the upper tail also extends from 7.89~kgCO$_2$ to 23.34~kgCO$_2$. This indicates that heavier charging demand makes low-emission scheduling progressively harder and increases variability across runs.}

    \textcolor{blue! }{Fig.\ref{fig:groupD_energy} shows the same monotonic pattern in operational metrics. Charged energy increases from 326.12~kWh to 523.68~kWh and 618.89~kWh, discharged energy rises from 106.91~kWh to 172.93~kWh, and transformer overload grows from 89.43~kWh to 180.54~kWh, with the largest uncertainty band in the high-demand case. Importantly, this is not a qualitative collapse of controller behaviour. Rather, the policy degrades smoothly and predictably as system stress increases, which supports the robustness of the proposed approach under changing demand conditions.}

    \textcolor{blue! }{Taken together, these supplementary experiments strengthen the main findings of the paper. The ablation study shows that each reward term serves a distinct function, the weight sensitivity study supports the choice of $W_{CO_{2}}=5$, the renewable-penetration analysis shows that the policy remains effective across a broad range of supply conditions, and the demand-variability analysis shows that performance degrades in a stable and interpretable manner as charging pressure increases. These results therefore provide direct evidence for the robustness and reliability of the proposed emission-aware RL controller.}

    \section{Conclusion}
        \label{conclusion}
    This paper employed the EV2Gym platform to evaluate heuristic, MPC, and RL controllers in realistic V2G smart charging scenarios. The RL-based strategy demonstrated strong adaptability across diverse renewable scenarios (Wind, Solar, and Hybrid), effectively balancing competing objectives such as minimising emissions, increasing renewable utilisation, and maintaining network constraints. RL achieved sustainability outcomes that matched or exceeded heuristic and MPC baselines while providing substantially faster inference during operation. Results were averaged over ten independent stochastic runs per scenario, ensuring statistical robustness.
    A key factor in RL performance was the emission-aware state representation combined with a multi-objective reward function. The reward penalised unmet charging demand and user dissatisfaction heavily (weight 50 each) while applying a smaller CO$_2$ penalty weight of 5 for charging during high CO$_2$ intensity periods. Although the RL agent did not explicitly optimise for average user satisfaction, it achieved final user satisfaction levels comparable to V2G-aware MPC, indicating that the learned policy implicitly preserved user charging needs.

The analysis clarified the role of MPC variants. G2V \textcolor{blue! }{prioritised} charging completion and achieved higher user satisfaction in most scenarios (approximately 62\% versus 42\% for V2G in the 50\% Wind case), while V2G reduced CO$_2$ emissions and improved renewable alignment. In the No-RE scenario, both variants converged to approximately 43\% satisfaction. These results illustrate a fundamental trade-off between user-centric fulfilment and system-level decarbonisation.

Scenarios with substantial wind contribution yielded higher renewable utilisation and lower emissions than solar-only cases. In the European context, wind provided broader temporal coverage beyond midday, supporting cleaner scheduling for both MPC-V2G and RL. Under high renewable penetration, the RL policy achieved up to 87\% emission reduction relative to uncontrolled baselines while maintaining grid constraints and acceptable user satisfaction.

The results demonstrate that an emission-aware RL strategy can simultaneously reduce CO$_2$ emissions, increase renewable \textcolor{blue! }{utilisation}, and respect grid constraints while maintaining acceptable user satisfaction and faster inference than MPC.

\textcolor{blue! }{From a deployment perspective, the trained SAC policy is lightweight at run time because online control reduces to a neural-network forward pass rather than a new optimisation solve at each step. The sub-millisecond inference time (0.116 ± 0.029 ms per step) makes the trained policy well-suited to both embedded EVSE deployment and cloud-side CPO execution~\cite{Damodarin2025smartEV}. Scalability remains the main challenge, since the action dimension grows directly with the number of EVSE ports. For larger fleets, this motivates shared-policy, hierarchical, or multi-agent decompositions instead of a single flat controller.~\cite{Keonwoo2022multi-agent} Integration with existing infrastructure is also feasible because Ireland already publishes real-time grid and carbon data through EirGrid, while the policy can be periodically retrained or site-specifically fine-tuned as fleet composition, charger ratings, and user behaviour evolve.~\cite{smartgrid-co2} In the Irish and wider European context, V2G rollout will also depend on interoperability and compliance, including smart-charging capability, standard EV–charger communication, backend protocol support, and local rules for connection, metering, and settlement of bidirectional energy flows.~\cite{eu2023afir, eu2025afir_amendment}}

\textcolor{blue! }{These deployment considerations should, however, be interpreted alongside several limitations. The present study is restricted to a single 24-hour horizon and does not examine multi-day or seasonal effects. It also assumes accurate forecast inputs for carbon intensity, while real deployments would face forecast error. In addition, the experiments consider one transformer and a fixed stochastic user profile, so broader network topologies and more extreme demand behaviours remain untested. Finally, although battery-aware simulation is included, battery degradation cost is not modelled as an explicit term in the RL reward.}

\textcolor{blue! }{Future work should build on these limitations in several directions. First, the controller should be extended from a single 24-hour horizon to multi-day and seasonal settings, potentially using memory-augmented RL to capture carry-over effects in user demand and renewable availability. Second, forecast uncertainty should be modelled explicitly, for example through robust or stochastic reward design, so that the policy remains reliable under imperfect carbon-intensity and renewable predictions. Third, the environment should be expanded from a single-transformer site to multi-transformer, multi-building, and feeder-level topologies. Fourth, multi-agent RL is a promising direction for decentralised coordination across chargers or sites as fleet size grows. Fifth, battery degradation cost should be included directly in the reward so that emissions, user satisfaction, and asset wear are optimised jointly. Finally, a real-world pilot with online fine-tuning would be valuable to validate practical integration, runtime performance, and policy adaptation under changing fleet composition and operating conditions.}

    \bibliographystyle{model1-num-names}

    \bibliography{ref}

@misc{IEA2024,
  author       = {{IEA}},
  title        = {Global EV Outlook 2024},
  year         = {2024},
  publisher    = {International Energy Agency},
  howpublished = {\url{https://www.iea.org/reports/global-ev-outlook-2024}},
  note         = {Licence: CC BY 4.0},
}

@article{rossi2025smart,
  title={Smart Electric Vehicle Charging Algorithm to Reduce the Impact on Power Grids: A Reinforcement Learning Based Methodology},
  author={Rossi, Federico and Diaz-Londono, Cesar and Li, Yang and Zou, Changfu and Gruosso, Giambattista},
  journal={IEEE Open Journal of Vehicular Technology},
  year={2025},
  publisher={IEEE}
}

@inproceedings{panda2024multi,
  title={A Multi-Objective Optimization Model for Smart EV Charging Scheduling Considering the Benefits of EV Owners and Distribution Network Operators},
  author={Panda, Subhadarshini and Ganguly, Sanjib},
  booktitle={2024 23rd National Power Systems Conference (NPSC)},
  pages={1--6},
  year={2024},
  organization={IEEE}
}

@techreport{NationalEVAnnualReport,
  title={National Electric Vehicle Infrastructure Formula Program Annual Report: Plan Year 2022-2023},
  author={Chu, Jean and Gilmore, Bridget and Hassol, Joshua and Jenn, Alan and Lommele, Steve and Myers, Lissa and Richardson, Heather and Schroeder, Alex and Shah, Monisha},
  year={2023},
  institution={National Renewable Energy Laboratory (NREL), Golden, CO (United States)}
}

@inproceedings{dan2024cooperative,
  title={A Cooperative Bargaining Game Framework for Vehicle-to-Vehicle Energy Sharing and Trading at Charging Stations},
  author={Dan, Mainak and Easwaran, Arvind},
  booktitle={2024 IEEE 27th International Conference on Intelligent Transportation Systems (ITSC)},
  pages={96--103},
  year={2024},
  organization={IEEE}
}

@article{9817043,
  title={A robust model predictive control-based scheduling approach for electric vehicle charging with photovoltaic systems},
  author={Yang, Yu and Yeh, Hen-Geul and Nguyen, Richard},
  journal={IEEE Systems Journal},
  volume={17},
  number={1},
  pages={111--121},
  year={2022},
  publisher={IEEE}
}

@misc{EV2Gym,
  title={EV2Gym: A Flexible V2G Simulator for EV Smart Charging Research and Benchmarking},
  author={Orfanoudakis, S and Diaz-Londono, C and Y{\i}lmaz, YE and Palensky, P and Vergara, PP},
  year={2024}
}

@article{10315215,
  title={Optimal electric vehicle charging strategies for long-distance driving},
  author={Liu, Bin and Ni, Wei and Liu, Ren Ping and Guo, Y Jay and Zhu, Hongbo},
  journal={IEEE Transactions on Vehicular Technology},
  volume={73},
  number={4},
  pages={4949--4960},
  year={2023},
  publisher={IEEE}
}

@misc{smartgrid-co2,
  author = "{EirGrid and SONI}",
  title = "{Smart Grid Dashboard: CO2 Intensity}",
  howpublished = "\url{https://www.smartgriddashboard.com/roi/co2/}",
  year = "2025"
}

@misc{pvgis-solar,
  author = "{European Commission Joint Research Centre}",
  title = "{Photovoltaic Geographical Information System (PVGIS)}",
  howpublished = "\url{https://re.jrc.ec.europa.eu/pvg_tools/en/}",
  year = "2025"
}

@article{LEE20236624,
  title={Vehicle-to-grid optimization considering battery aging},
  author={Lee, Chih Feng and Bjurek, Kalle and Hagman, Victor and Li, Yang and Zou, Changfu},
  journal={IFAC-PapersOnLine},
  volume={56},
  number={2},
  pages={6624--6629},
  year={2023},
  publisher={Elsevier}
}

@article{SAXENA2015720,
title = {Charging ahead on the transition to electric vehicles with standard 120V wall outlets},
journal = {Applied Energy},
volume = {157},
pages = {720-728},
year = {2015},
issn = {0306-2619},
doi = {https://doi.org/10.1016/j.apenergy.2015.05.005},
url = {https://www.sciencedirect.com/science/article/pii/S0306261915005899},
author = {Samveg Saxena and Jason MacDonald and Scott Moura}
}

@inproceedings{1709436,
  title={Integration of wind power generation in the Irish grid},
  author={Dudurych, IM and Holly, M and Power, M},
  booktitle={2006 IEEE Power Engineering Society General Meeting},
  pages={8--pp},
  year={2006},
  organization={IEEE}
}

@article{9246271,
  title={System impact studies for near 100\% renewable energy systems dominated by inverter based variable generation},
  author={Holttinen, Hannele and Kiviluoma, Juha and Flynn, Damian and Smith, J Charles and Orths, Antje and Eriksen, Peter B{\o}rre and Cutululis, Nicolaos and S{\"o}der, Lennart and Korp{\aa}s, Magnus and Estanqueiro, Ana and others},
  journal={IEEE Transactions on Power Systems},
  volume={37},
  number={4},
  pages={3249--3258},
  year={2020},
  publisher={IEEE}
}

@inproceedings{10253224,
  title={Analysis of wind energy curtailment in the ireland and northern ireland power systems},
  author={Hurtado, Manuel and K{\"e}r{\c{c}}i, Taulant and Tweed, Simon and Kennedy, Eoin and Kamaluddin, Nezar and Milano, Federico},
  booktitle={2023 IEEE Power \& Energy Society General Meeting (PESGM)},
  pages={1--5},
  year={2023},
  organization={IEEE}
}

@article{GANESH2022111833,
  title={A review of reinforcement learning based energy management systems for electrified powertrains: Progress, challenge, and potential solution},
  author={Ganesh, Akhil Hannegudda and Xu, Bin},
  journal={Renewable and Sustainable Energy Reviews},
  volume={154},
  pages={111833},
  year={2022},
  publisher={Elsevier}
}

@inproceedings{10688856,
  title={Safety-Aware Reinforcement Learning for Electric Vehicle Charging Station Management in Distribution Network},
  author={Fan, Jiarong and Liebman, Ariel and Wang, Hao},
  booktitle={2024 IEEE Power \& Energy Society General Meeting (PESGM)},
  pages={1--5},
  year={2024},
  organization={IEEE}
}

@article{DIAZLONDONO2024100326,
  title={Enhanced EV charging algorithm considering data-driven workplace chargers categorization with multiple vehicle types},
  author={Diaz-Londono, Cesar and Fambri, Gabriele and Maffezzoni, Paolo and Gruosso, Giambattista},
  journal={eTransportation},
  volume={20},
  pages={100326},
  year={2024},
  publisher={Elsevier}
}

@article{QIU2023113052,
  title={Reinforcement learning for electric vehicle applications in power systems: A critical review},
  author={Qiu, Dawei and Wang, Yi and Hua, Weiqi and Strbac, Goran},
  journal={Renewable and Sustainable Energy Reviews},
  volume={173},
  pages={113052},
  year={2023},
  publisher={Elsevier}
}

@article{8909765,
  title={2019 IEEE International Conference on Communications, Control, and Computing Technologies for Smart Grids (SmartGridComm)},
  author={Zhang, F and Wang, Z and Li, Y and Zhang, C},
  year={2019},
  publisher={IEEE Piscataway}
}

@article{NEURIPS2023_ba748557,
  title={SustainGym: Reinforcement learning environments for sustainable energy systems},
  author={Yeh, Christopher and Li, Victor and Datta, Rajeev and Arroyo, Julio and Christianson, Nicolas and Zhang, Chi and Chen, Yize and Hosseini, Mohammad Mehdi and Golmohammadi, Azarang and Shi, Yuanyuan and others},
  journal={Advances in Neural Information Processing Systems},
  volume={36},
  pages={59464--59476},
  year={2023}
}

@article{8710609,
  title={Offline and online electric vehicle charging scheduling with V2V energy transfer},
  author={Koufakis, Alexandros-Michail and Rigas, Emmanouil S and Bassiliades, Nick and Ramchurn, Sarvapali D},
  journal={IEEE Transactions on Intelligent Transportation Systems},
  volume={21},
  number={5},
  pages={2128--2138},
  year={2019},
  publisher={IEEE}
}

@inproceedings{9242538,
  title={Optimal business case for provision of grid services through EVs with V2G capabilities},
  author={Meenakumar, Pragadeesh and Aunedi, Marko and Strbac, Goran},
  booktitle={2020 Fifteenth International Conference on Ecological Vehicles and Renewable Energies (EVER)},
  pages={1--10},
  year={2020},
  organization={IEEE}
}

@article{RIGAS201899,
  title={EVLibSim: A tool for the simulation of electric vehicles’ charging stations using the EVLib library},
  author={Rigas, Emmanouil S and Karapostolakis, Sotiris and Bassiliades, Nick and Ramchurn, Sarvapali D},
  journal={Simulation Modelling Practice and Theory},
  volume={87},
  pages={99--119},
  year={2018},
  publisher={Elsevier}
}

@article{MORSTYN2020115397,
  title={OPEN: An open-source platform for developing smart local energy system applications},
  author={Morstyn, Thomas and Collett, Katherine A and Vijay, Avinash and Deakin, Matthew and Wheeler, Scot and Bhagavathy, Sivapriya M and Fele, Filiberto and McCulloch, Malcolm D},
  journal={Applied Energy},
  volume={275},
  pages={115397},
  year={2020},
  publisher={Elsevier}
}

@inproceedings{karatzinis2022chargym,
  title={Chargym: An ev charging station model for controller benchmarking},
  author={Karatzinis, Georgios and Korkas, Christos and Terzopoulos, Michalis and Tsaknakis, Christos and Stefanopoulou, Aliki and Michailidis, Iakovos and Kosmatopoulos, Elias},
  booktitle={IFIP International Conference on Artificial Intelligence Applications and Innovations},
  pages={241--252},
  year={2022},
  organization={Springer}
}

@article{holkar2010overview,
  title={An overview of model predictive control},
  author={Holkar, Kailas S and Waghmare, Laxman M},
  journal={International Journal of control and automation},
  volume={3},
  number={4},
  pages={47--63},
  year={2010}
}

@article{li2017deep,
  title={Deep reinforcement learning: An overview},
  author={Li, Yuxi},
  journal={arXiv preprint arXiv:1701.07274},
  year={2017}
}

@article{kempton2005vehicle,
  title={Vehicle-to-grid power fundamentals: Calculating capacity and net revenue},
  author={Kempton, Willett and Tomi{\'c}, Jasna},
  journal={Journal of power sources},
  volume={144},
  number={1},
  pages={268--279},
  year={2005},
  publisher={Elsevier}
}

@article{csengor2018optimal,
  title={Optimal energy management of EV parking lots under peak load reduction based DR programs considering uncertainty},
  author={{\c{S}}eng{\"o}r, {\.I}brahim and Erdin{\c{c}}, Ozan and Yener, Bar{\i}{\c{s}} and Ta{\c{s}}c{\i}karao{\u{g}}lu, Ak{\i}n and Catalao, Joao PS},
  journal={IEEE Transactions on Sustainable Energy},
  volume={10},
  number={3},
  pages={1034--1043},
  year={2018},
  publisher={IEEE}
}

@article{raffin2021stable,
  title={Stable-baselines3: Reliable reinforcement learning implementations},
  author={Raffin, Antonin and Hill, Ashley and Gleave, Adam and Kanervisto, Anssi and Ernestus, Maximilian and Dormann, Noah},
  journal={Journal of machine learning research},
  volume={22},
  number={268},
  pages={1--8},
  year={2021}
}

@article{qiu2024graph,
  title={Graph reinforcement learning for carbon-aware electric vehicles in power-transport networks},
  author={Qiu, Dawei and Wang, Yi and Ding, Zhaohao and Strbac, Goran},
  journal={IEEE Transactions on Smart Grid},
  volume={15},
  number={4},
  pages={3919--3935},
  year={2024},
  publisher={IEEE}
}

@inproceedings{yilmaz2024reinforcement,
  title={Reinforcement learning for optimized EV charging through power setpoint tracking},
  author={Y{\i}lmaz, Yunus Emre and Orfanoudakis, Stavros and Vergara, Pedro P},
  booktitle={2024 IEEE PES Innovative Smart Grid Technologies Europe (ISGT EUROPE)},
  pages={1--5},
  year={2024},
  organization={IEEE}
}

@article{qureshi2024multiobjective,
  title={Multiobjective pareto-optimal intelligent electric vehicle charging schedule in a commercial charging station: A Stochastic convex optimization approach},
  author={Qureshi, Ubaid and Ghosh, Arnob and Panigrahi, Bijaya Ketan},
  journal={IEEE Transactions on Industrial Informatics},
  year={2024},
  publisher={IEEE}
}

@article{hosseini2024optimizing,
  title={Optimizing electric vehicle charging through an artificial intelligence mechanism for smart transportation},
  author={Hosseini, Samira and Yassine, Abdulsalam and Hossain, M Shamim},
  journal={IEEE Internet of Things Journal},
  year={2024},
  publisher={IEEE}
}

@inproceedings{kanchana2024optimizing,
  title={Optimizing Electric Vehicle Charging Schedules: Minimizing Charging Time and Travel Distance},
  author={Kanchana, Wutthipum and Singh, Jai Govind and Ongsakul, Weerakorn and others},
  booktitle={2024 International Conference on Sustainable Energy: Energy Transition and Net-Zero Climate Future (ICUE)},
  pages={1--5},
  year={2024},
  organization={IEEE}
}

@article{Damodarin2025smartEV,
  author  = {Damodarin, Udhaya and Cardarilli, Gian Carlo and Di Nunzio, Luca and Re, Marco and Span{\`o}, Sergio},
  title   = {Smart Electric Vehicle Charging Management Using Reinforcement Learning on {FPGA} Platforms},
  journal = {Sensors},
  year    = {2025},
  volume  = {25},
  number  = {8},
  pages   = {2585},
  doi     = {10.3390/s25082585}
}

@article{Keonwoo2022multi-agent,
  title   = {Multi-agent deep reinforcement learning approach for EV charging scheduling in a smart grid},
  journal = {Applied Energy},
  volume  = {328},
  pages   = {120111},
  year    = {2022},
  author  = {Keonwoo Park and Ilkyeong Moon},
  doi     = {10.1016/j.apenergy.2022.120111}
}

@misc{eu2023afir,
  title        = {Regulation (EU) 2023/1804 of the European Parliament and of the Council of 13 September 2023 on the deployment of alternative fuels infrastructure, and repealing Directive 2014/94/EU},
  author       = {{European Parliament and the Council of the European Union}},
  year         = {2023},
  month        = sep,
  journal      = {Official Journal of the European Union},
  volume       = {L 234},
  pages        = {1--47},
  url          = {http://data.europa.eu/eli/reg/2023/1804/oj},
  note         = {Text with EEA relevance}
}

@misc{eu2025afir_amendment,
  title        = {Commission Delegated Regulation (EU) 2025/656 of 2 April 2025 amending Regulation (EU) 2023/1804 of the European Parliament and of the Council as regards standards for wireless recharging, electric road system, vehicle-to-grid communication and hydrogen supply for road transport vehicles},
  author       = {{European Commission}},
  year         = {2025},
  month        = jun,
  journal      = {Official Journal of the European Union},
  number       = {2025/656},
  pages        = {1--10},
  url          = {http://data.europa.eu/eli/reg_del/2025/656/oj},
  note         = {Published in the Official Journal of the European Union, 18 June 2025}
}

@inproceedings{fan2024safety,
  author    = {Jiarong Fan and Ariel Liebman and Hao Wang},
  title     = {Safety-Aware Reinforcement Learning for Electric Vehicle Charging Station Management in Distribution Network},
  booktitle = {2024 IEEE Power and Energy Society General Meeting (PESGM)},
  year      = {2024},
  publisher = {IEEE},
  doi       = {10.1109/PESGM51994.2024.10688856}
}

@article{zhao2025maasac,
  author  = {Qiang Zhao and Chengwei Xu and Chuan Sun and Yinghua Han},
  title   = {Smart residential electric vehicle charging and discharging scheduling via multi-agent asynchronous-updating deep reinforcement learning},
  journal = {Computers and Electrical Engineering},
  volume  = {126},
  pages   = {110473},
  year    = {2025},
  doi     = {10.1016/j.compeleceng.2025.110473}
}

@article{silva2025carbon,
  author  = {Carlos A. M. Silva and Ricardo J. Bessa},
  title   = {Carbon-aware dynamic tariff design for electric vehicle charging stations with explainable stochastic optimization},
  journal = {Applied Energy},
  volume  = {389},
  pages   = {125674},
  year    = {2025},
  doi     = {10.1016/j.apenergy.2025.125674}
}

@book{ipcc2006guidelines,
  author    = {{Intergovernmental Panel on Climate Change (IPCC)}},
  title     = {2006 {IPCC} Guidelines for National Greenhouse Gas Inventories},
  year      = {2006},
  publisher = {Institute for Global Environmental Strategies (IGES)},
  address   = {Hayama, Japan},
  editor    = {Simon Eggleston and Leandro Buendia and Kyoko Miwa and Todd Ngara and Kiyoto Tanabe},
  note      = {Prepared by the National Greenhouse Gas Inventories Programme},
  url       = {https://www.ipcc-nggip.iges.or.jp/public/2006gl/}
}

@book{ghgprotocol2015scope2,
  author    = {{World Resources Institute (WRI)} and {World Business Council for Sustainable Development (WBCSD)}},
  title     = {{GHG} Protocol Scope 2 Guidance: An Amendment to the {GHG} Protocol Corporate Standard},
  year      = {2015},
  publisher = {World Resources Institute},
  address   = {Washington, DC},
  url       = {https://ghgprotocol.org/sites/default/files/2023-03/Scope%202%20Guidance.pdf}
}


    %



\end{document}